\documentclass[letterpaper]{article} 
\usepackage{aaai2027}  
\usepackage[hyphens]{url}  
\usepackage{graphicx} 
\usepackage{natbib}  
\usepackage{caption} 
\usepackage{algorithm}
\usepackage{algorithmic}
\usepackage{amssymb}
\usepackage{amsmath}

\usepackage{enumitem}
\usepackage{dirtree}
\usepackage{amsfonts}
\usepackage{pifont}

\newcommand{\cmark}{\ding{51}}
\newcommand{\xmark}{\ding{55}}

\usepackage{newfloat}
\usepackage{listings}
\DeclareCaptionStyle{ruled}{labelfont=normalfont,labelsep=colon,strut=off} 
\floatstyle{ruled}
\newfloat{listing}{tb}{lst}{}
\floatname{listing}{Listing}

\usepackage{booktabs}

\title{Any3DAvatar: Fast and High-Quality Full-Head 3D Avatar Reconstruction from a Single Portrait Image}
\author{
   Yujie Gao\textsuperscript{\rm 1}\equalcontrib, Yao Xiao\textsuperscript{\rm 1}\equalcontrib, Xiangnan Zhu\textsuperscript{\rm 1}, Ya Li\textsuperscript{\rm 1}, Yiyi Zhang\textsuperscript{\rm 1}, Liqing Zhang\textsuperscript{\rm 1}, Jianfu Zhang\textsuperscript{\rm 1}\corresponding
}
\affiliations{
    \textsuperscript{\rm 1}Shanghai Jiao Tong University,\\

    \{GYjgyj123, xiaoyao0517, Zxn\_049, 696969, yi95yi, zhang-lq, c.sis\}@sjtu.edu.cn;

}

\begin{document}

\twocolumn[{
\renewcommand\twocolumn[1][]{#1}
\maketitle
\begin{center}
    \centering
    \captionsetup{type=figure,skip=10pt}
    \includegraphics[width=0.85\textwidth]{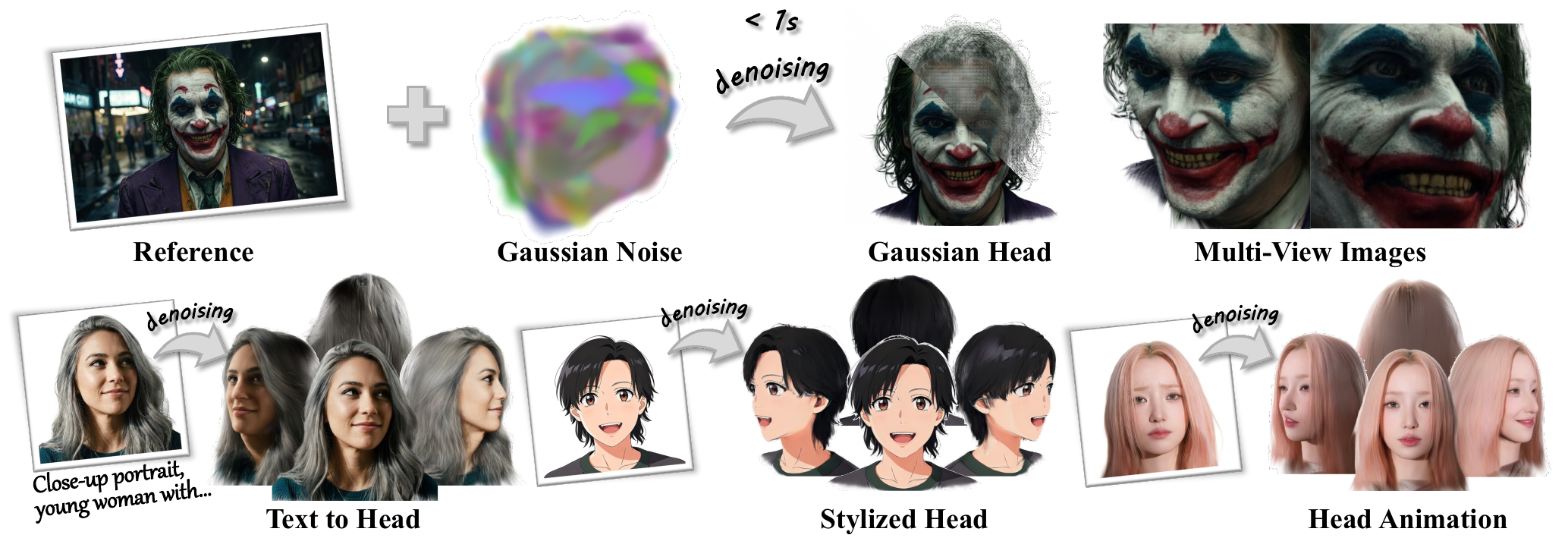}
    \captionof{figure}{Any3DAvatar generates high-quality 3D heads in under one second and supports diverse downstream applications.}
\end{center}
}]

\begin{abstract}
This paper addresses single-image full-head 3D avatar reconstruction, where existing methods remain limited by two key bottlenecks: Existing approaches often exhibit limitations in teeth reconstruction, accessory modeling, complete-head geometry recovery, and robustness to non-canonical inputs, while high-quality methods commonly rely on heavy preprocessing and costly per-subject optimization.
To solve these problems, we propose Any3DAvatar, a single-stage 3D Gaussian head reconstruction framework that directly generates a complete 3D head from a single portrait through few-step conditional denoising.
We further introduce auxiliary view-conditioned appearance supervision to preserve novel-view textures at no additional inference cost.
We also construct AnyHead, a data suite with diverse identities, dense multi-view supervision, and accessory-rich samples for robust training.
Extensive evaluations on public datasets against recent full-head reconstruction baselines demonstrate that Any3DAvatar achieves the best overall rendering fidelity and identity consistency, with inference time as low as 0.64 seconds in its fastest setting.
\end{abstract}

\begin{links}
    \link{Project Page}{https://biggaoga.github.io/Any3DAvatar/}
\end{links}

\section{Introduction}

Single-Image full-head 3D reconstruction aims to reconstruct a complete, identity-consistent, and controllable 3D head representation from only one RGB portrait, including challenging regions such as teeth, side profiles, and the back head that are not directly observed in the input view. This task is important not only because it supports practical applications in virtual/augmented reality, gaming, and digital content production, but also because it serves as a foundation for downstream tasks such as talking-head synthesis, 3D head editing and virtual try-on. Consequently, improving both reconstruction fidelity and inference efficiency is essential for real-world deployment.

Despite recent progress, existing methods still face two key bottlenecks: reconstruction quality and inference efficiency.
\textbf{(1) Quality.} Existing methods often struggle to preserve full-head fidelity on unconstrained real portraits, especially in challenging regions such as hair, side/back head, accessories, and teeth. 3DMM- or mesh-prior-based methods~\cite{3DMM, DECA, MICA, 3DDFA_v2, 3DDFA_v3, pixel3dmm, gpavatar, gagavatar, cap4d, lam, FLAME} are limited by predefined head topology or front-face bias, while implicit or two-stage approaches~\cite{panohead, spherehead,rodinhd, facelift, hqhead} can suffer from weak cross-view consistency. Moreover, existing datasets~\cite{nersemble, ava-256, renderme360, cafca} suffer from limited identity diversity, sparse viewpoint coverage, and insufficient accessory representation, thereby weakening model generalization.
\textbf{(2) Efficiency.} Practical deployment is also limited by heavy preprocessing, and costly per-subject optimization. Many methods require head fitting before reconstruction~\cite{DECA, gpavatar, gagavatar, cap4d, lam}, while recent high-quality approaches still rely on per-subject optimization, multi-view fitting, or SDS-style objectives~\cite{panohead, id_sculpt, hqhead, arc2avatar}. As a result, inference can take minutes or even hours per subject, limiting scalability in real applications.

To address these limitations, we propose \textbf{Any3DAvatar}, a unified framework for fast and high-quality full 3D head reconstruction from a single portrait, with three coordinated contributions.
\textbf{(1) AnyHead Dataset.} To mitigate dataset bias (limited diversity, sparse viewpoints, and weak accessory coverage), we construct AnyHead with a large diverse identity pool, dense and controllable multi-view supervision, and an accessory-rich subset, so that training directly covers hard cases such as large pose changes, side/back regions, and wearable items.
\textbf{(2) Few-Step Denoising Network.} Leveraging the strong structural prior of human heads, we design a DiT~\cite{DiT}-based single-stage few-step denoising network that starts from a Plücker-conditioned multi-view noise-token initialization and reconstructs full-head 3D outputs without per-subject optimization. Our ablations show that approximately five denoising steps yield the best reconstruction quality, whereas one-step inference achieves a runtime of only 0.64 seconds with a minor quality drop, enabling a flexible trade-off between quality and efficiency.
\textbf{(3) Auxiliary View-Conditioned Appearance Supervision.} To address novel-view detail degradation, we introduce auxiliary view-conditioned appearance supervision with a VAE decoder~\cite{vae} alongside Gaussian reconstruction, improving novel-view appearance and texture fidelity at no additional inference cost.

The contributions of Any3DAvatar are summarized as follows:
\begin{itemize}
    \item We construct AnyHead, a high-quality and diverse 3D head dataset that broadens training coverage across identities, viewpoints, and accessories.
    \item We propose a single-stage denoising network for single-image full 3D head reconstruction, achieving faster few-step inference while maintaining high quality.
    \item We introduce auxiliary view-conditioned appearance supervision alongside Gaussian reconstruction, improving novel-view details with zero extra inference burden.
\end{itemize}

\section{Related Works}
\noindent\textbf{Multi-View Generation Methods:}
For single-image or text conditioned 3D reconstruction, generating multi-view images has long been a key intermediate step that bridges sparse input observations and geometry-complete 3D modeling. It provides richer cross-view appearance cues and more stable geometric constraints before final 3D reconstruction. Representative methods in this line include MVDream~\cite{mvdream}, Zero-1-to-3~\cite{zero123}, Zero123++~\cite{zero123++}, Unique3D~\cite{unique3d}, Wonder3D~\cite{wonder3d}, and SVD-based pipelines~\cite{svd, sv3d}. However, as direct single-image/text-to-3D generation continues to advance rapidly, this intermediate multi-view generation line is gradually becoming less central in the field.

\noindent\textbf{3D Reconstruction Methods:}
The recent task of 3D reconstruction is to recover a 3D representation from other modalities (e.g., text, single-view images, or multi-view images). According to the input content and the reconstruction strategy, existing methods can be roughly grouped into two categories. One category takes unconstrained multi-view images as input and directly predicts camera parameters together with reconstruction outputs (e.g., point clouds). Typical examples are $\pi^3$~\cite{pi3}, VGGT~\cite{vggt}, Fast3R ~\cite{fast3r}, and DUSt3R~\cite{dust3r}. The other category directly decodes 3D representations from a single image or text with diffusion-based models, where the output representation can be 3DGS, mesh, point cloud, or voxel; representative works include OpenDiffusionGS~\cite{diffusiongs} and TRELLIS~\cite{trellis}.
As for 3D representation, 3D Gaussian Splatting (3DGS)~\cite{3dgs} has become a widely used one for high-quality and efficient 3D reconstruction in recent years. By modeling scenes with anisotropic Gaussians and differentiable splatting, 3DGS provides a strong quality-efficiency trade-off for head reconstruction, preserving fine details while supporting real-time rendering, especially for semi-transparent regions such as hair.


\noindent\textbf{3D Head Reconstruction from Single Images:}
Compared with multi-view reconstruction, single-view 3D head reconstruction is substantially more ill-posed due to severe depth and occlusion ambiguities. Early methods ~\cite{DECA, pixel3dmm, MICA, 3DDFA_v2, 3DDFA_v3} are typically built on statistical head models such as 3DMM ~\cite{3DMM} and FLAME ~\cite{FLAME}, which map real facial landmarks or image cues to a canonical parametric head space.
Recent single-view methods further evolve into two major directions. One line directly generates dynamic 4D heads from monocular image/video inputs. These methods~\cite{gpavatar, gagavatar, cap4d, lam, fastavatar,OMEGA-Avatar2026} often achieve strong temporal consistency, but their per-frame static quality is usually limited, and the synthesized viewpoints are often constrained to narrow facial regions. The other line focuses on directly producing high-quality full 3D heads from single-view inputs. These methods~\cite{panohead, spherehead, id_sculpt, hqhead, arc2avatar, diffportrait360, panoLAM, PercHead} generally provide better static fidelity and fuller head coverage, yet they may suffer from consistency issues when extended to continuous video-space generation.

\section{Methodology}

\begin{figure*}[t]
  \centering
  \includegraphics[width=0.89\textwidth]{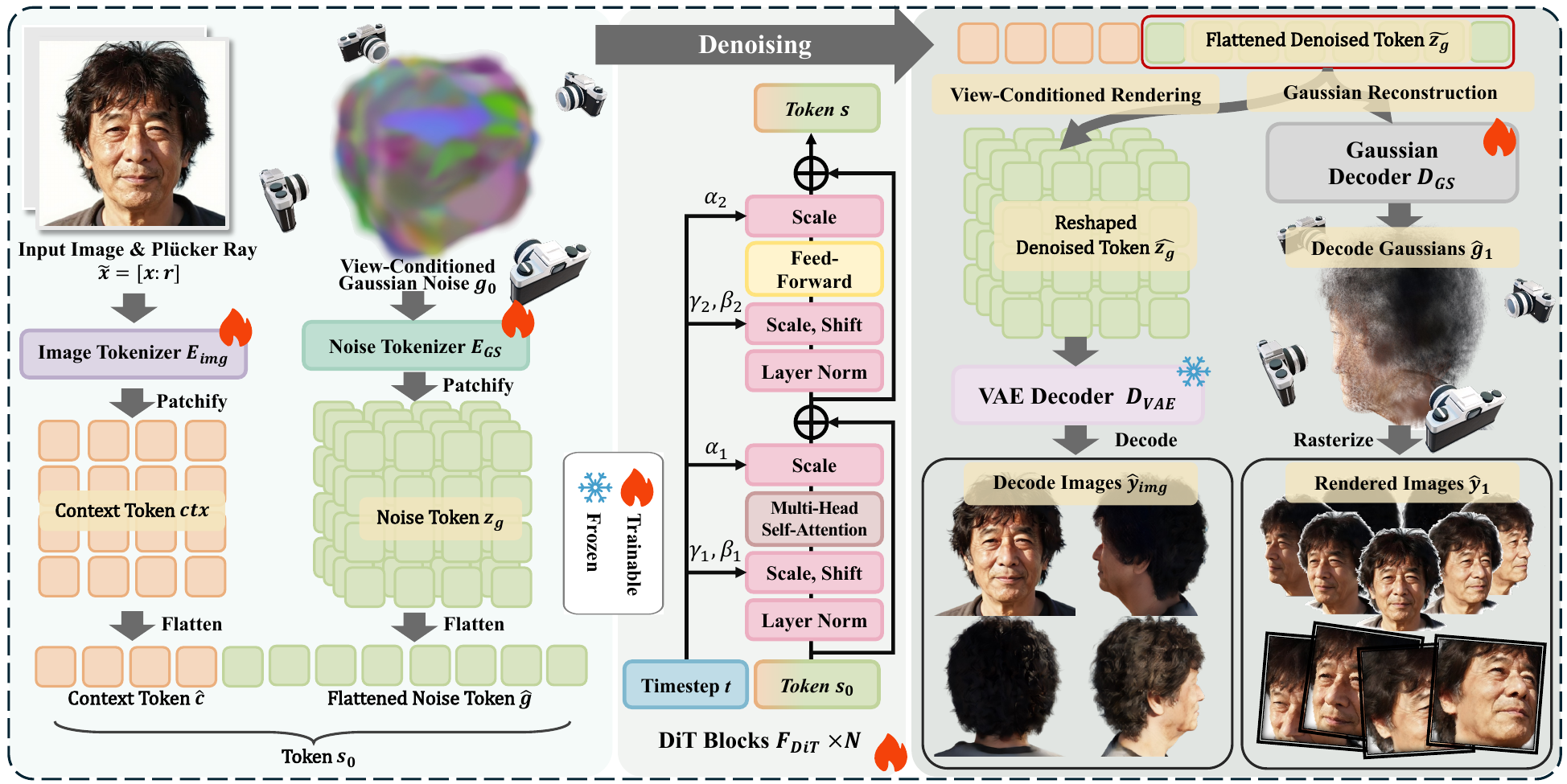}
  \caption{\textbf{Pipeline of Any3DAvatar.} Given a single portrait, we extract image and Gaussian tokens, jointly denoise them with a DiT backbone, and decode the outputs into 3D Gaussian point clouds through a few-step feed-forward path at inference. During training, we additionally supervise the same tokens with a view-conditioned image decoding branch.}
  \label{fig:pipeline}
\end{figure*}

\subsection{Problem Formulation}
Given an identity $\mathcal{I}$ and a single RGB portrait $\mathbf{x} \in \mathbb{R}^{3 \times H \times W}$ (frontal view by default), a Pl\"ucker ray embedding $\mathbf{r} \in \mathbb{R}^{6 \times H \times W}$ is computed and concatenated with the image to form a spatially aware input tensor $\tilde{\mathbf{x}}=[\mathbf{x};\mathbf{r}] \in \mathbb{R}^{9 \times H \times W}$. Let the patch size be $p \in \mathbb{N}^{+}$ with $p \mid H$ and $p \mid W$. The tensor $\tilde{\mathbf{x}}$ is partitioned into non-overlapping $p \times p$ patches and encoded by an image tokenizer $E_{\text{img}}(\cdot)$, producing context features $\mathbf{ctx}=E_{\text{img}}(\tilde{\mathbf{x}}) \in \mathbb{R}^{N_h \times N_w \times 1024}$, where $N_h=\lfloor H/p 
\rfloor$ and $N_w=\lfloor W/p 
\rfloor$.
Four canonical viewpoints are defined as $\mathcal{V}=\{\text{front},\text{left},\text{right},\text{back}\}$, with corresponding Pl\"ucker rays $\mathbf{R} \in \mathbb{R}^{4 \times H \times W \times 6}$. For each of the four canonical views, we construct Gaussian noise $\mathbf{g}_{0}$ by concatenating 3-channel image noise with the corresponding Pl\"ucker ray embedding, yielding $\mathbf{g}_{0} \in \mathbb{R}^{4 \times H \times W \times 9}$. Using the same patch size $p$, view-wise Gaussian features are tokenized and encoded by a Gaussian tokenizer $E_{\text{GS}}(\cdot)$ to obtain $\mathbf{z}_g=E_{\text{GS}}(\mathbf{g}_0) \in \mathbb{R}^{4 \times N_h \times N_w \times 1024}$.
Spatial dimensions are then flattened in both branches: $\hat{\mathbf{c}}=\mathrm{Flatten}(\mathbf{ctx}) \in \mathbb{R}^{(N_hN_w) \times 1024}$ and $\hat{\mathbf{g}}=\mathrm{Flatten}(\mathbf{z}_g) \in \mathbb{R}^{(4N_hN_w) \times 1024}$. Concatenating them along the token dimension yields the DiT input sequence $\mathbf{s}_0=[\hat{\mathbf{c}};\hat{\mathbf{g}}] \in \mathbb{R}^{(5N_hN_w) \times 1024}$, which is processed by $F_{\text{DiT}}(\cdot)$ as $\mathbf{s}=F_{\text{DiT}}(\mathbf{s}_0)$. The denoised Gaussian tokens $\tilde{\mathbf{z}}_g$ are split from $\mathbf{s}$.
Finally, two decoding heads are applied: the VAE image decoder $D_{\text{vae}}(\cdot)$ from Stable Diffusion 2.1, which first reshapes tokens $\tilde{\mathbf{z}}_g$ to $\hat{\mathbf{z}}_g$ and predicts four-view images $\hat{\mathbf{y}}_{img}=D_{\text{vae}}(\hat{\mathbf{z}}_g) \in \mathbb{R}^{4 \times H \times W \times 3}$, and a Gaussian decoder $D_{\text{gs}}(\cdot)$ that predicts Gaussian parameters $\mathbf{\hat{g}}_{1}=D_{\text{gs}}(\tilde{\mathbf{z}}_g) \in \mathbb{R}^{(4HW) \times 14}$, which are then assembled into a Gaussian point cloud and rendered to obtain multi-view images $\hat{\mathbf{y}}_{1}$.

\subsection{AnyHead Dataset}
\begin{figure*}[t]
  \centering
  \includegraphics[width=1.0\textwidth]{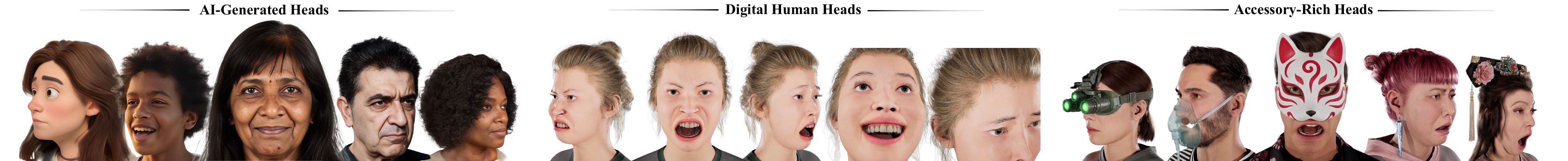}
  \caption{Overview of the three complementary subsets of our AnyHead dataset.}
  \label{fig:dataset}
\end{figure*}

To support robust few-step generation under diverse real-world conditions, we construct \textbf{AnyHead} (Fig.~\ref{fig:dataset}), a three-part dataset suite that jointly improves data diversity, geometric supervision, and appearance realism. Specifically, AnyHead consists of AI-Generated Heads, Digital Human Heads, and Accessory-Rich Heads.

\noindent\textbf{AI-Generated Heads:}
Real-captured multi-view head datasets often suffer from selection bias and limited identity coverage, which can lead to distribution shift at test time. To mitigate this issue, we construct an AI-generated subset with nearly 1,000 identities and broad attribute diversity. Our pipeline has three stages: (i) uniformly sampling identity-related attributes (e.g., age, ethnicity, hairstyle, expression, and lighting), (ii) generating frontal portraits with a text-to-image model ~\cite{z-image} and converting them into 3D Gaussian heads via an image-to-3D model ~\cite{facelift} for multi-view rendering, where the selected model provides strong head-shape and appearance diversity among existing 3D head reconstruction pipelines, and (iii) enhancing image quality with a face-detail enhancement model ~\cite{codeformer} and filtering low-quality samples using automatic identity/view-consistency checks. This subset improves identity coverage and reduces sampling bias for robust training.

\noindent\textbf{Digital Human Heads:}
At the same time, real-captured datasets still provide limited viewpoint coverage due to constrained camera quantity and placement. To address this limitation, we use Unreal Engine (UE) to synthesize high-fidelity digital human heads with dense, controllable camera sampling. We first sample real head keypoints from FaceScape ~\cite{facescape}, and then construct multi-expression digital heads conditioned on these keypoints. Our dataset contains 79 identities, each with 24 expressions. We adopt two complementary view-sampling strategies: (i) fixed circular views, where 40 cameras are placed at equal distance around the head across multiple elevation angles to provide global geometric supervision, and (ii) random views, where 40 cameras are sampled with random positions and orientations to emphasize local regions and fine-grained details. Together, these strategies provide both complete geometric coverage and strong local-detail supervision.

\noindent\textbf{Accessory-Rich Heads:}
Existing head datasets often underrepresent accessories. To bridge this gap, we construct an accessory-rich multi-view subset through controlled accessory augmentation of Digital Human Heads. The subset covers 95 accessory types, with 4 variants generated for each of 79 identities. To construct these samples, we arrange sampled multi-view images into a single multi-view grid and edit the grid with Nano Banana Pro to add accessories (e.g., glasses, hats, and other wearable items). We then manually remove samples with noticeable cross-view inconsistency in identity, geometry, or accessory appearance, and finally split each edited grid back into individual views to recover high-quality multi-view tuples.

Overall, AnyHead combines identity diversity, dense viewpoint coverage, and accessory realism to provide a stronger training basis for single-image 3D head avatar reconstruction. See the supplementary material for more dataset details.

\subsection{Few-Step Generation Network}
We introduce a DiT-based feed-forward generator that reconstructs a full-head 3D Gaussian representation from a single portrait. Benefiting from the strong structural priors inherent in human heads, the generator achieves high-quality reconstruction with only a few denoising steps.

\noindent\textbf{Network Design:}
As shown in Fig.~\ref{fig:pipeline}, the portrait and its Pl\"ucker ray embeddings are tokenized as context, while noisy multi-view images and their ray embeddings are encoded as denoising tokens. Both tokenizers use patch partitioning and MLP projection. An AdaLN-based DiT~\cite{DiT}, conditioned on timestep $t$, predicts denoised tokens that feed two heads: a VAE decoder for view-conditioned image reconstruction and a Gaussian decoder for 3D Gaussian prediction. The latter upsamples the tokens and predicts position, scale, rotation, opacity, and color.

\noindent\textbf{Denoising Formulation:}
Since ground-truth Gaussian point clouds are unavailable, we formulate conditional denoising in the rendered multi-view image space. Given clean multi-view images $\mathbf{y}_{1}$ and image noise $\mathbf{y}_{0}$, we sample $t\sim\mathcal{U}(0,1)$ and construct
\begin{equation}
\mathbf{y}_{t}=(1-t)\mathbf{y}_{0}+t\mathbf{y}_{1}.
\end{equation}
We concatenate $\mathbf{y}_{t}$ with its Pl\"ucker ray embeddings and tokenize it as $\mathbf{g}_{t}$. Conditioned on portrait $\tilde{\mathbf{x}}$, the DiT predicts tokens that are decoded into 3D Gaussians and rendered into multi-view images:
\begin{equation}
\tilde{\mathbf{z}}_{g}=F_{\theta}(\mathbf{g}_{t}, t, \tilde{\mathbf{x}}), \quad
\hat{\mathbf{g}}_1=D_{\mathrm{gs}}(\tilde{\mathbf{z}}_{g}), \quad
\hat{\mathbf{y}}_{1}=\mathcal{R}(\hat{\mathbf{g}}_1, \Pi),
\end{equation}
where $F_{\theta}$, $D_{\mathrm{gs}}$, and $\mathcal{R}$ denote the DiT denoiser, Gaussian decoder, and differentiable renderer, respectively, and $\Pi$ denotes the cameras. We define the predicted displacement and its target in the rendered image space as
\begin{equation}
\hat{\mathbf{d}}_{t}=\hat{\mathbf{y}}_{1}-\mathbf{y}_{t}, \quad
\mathbf{d}_{t}^{\mathrm{gt}}=\mathbf{y}_{1}-\mathbf{y}_{t}.
\end{equation}
The Gaussian generation branch is trained with
\begin{equation}
\mathcal{L}_{\mathrm{denoise}}=\left\|\hat{\mathbf{d}}_{t}-\mathbf{d}_{t}^{\mathrm{gt}}\right\|_{2}^{2}.
\end{equation}
This displacement-based objective avoids division by $1-t$ and reduces sensitivity to timestep sampling, particularly near $t=1$.


\begin{figure*}[t]
  \centering
  \includegraphics[width=0.98\textwidth]{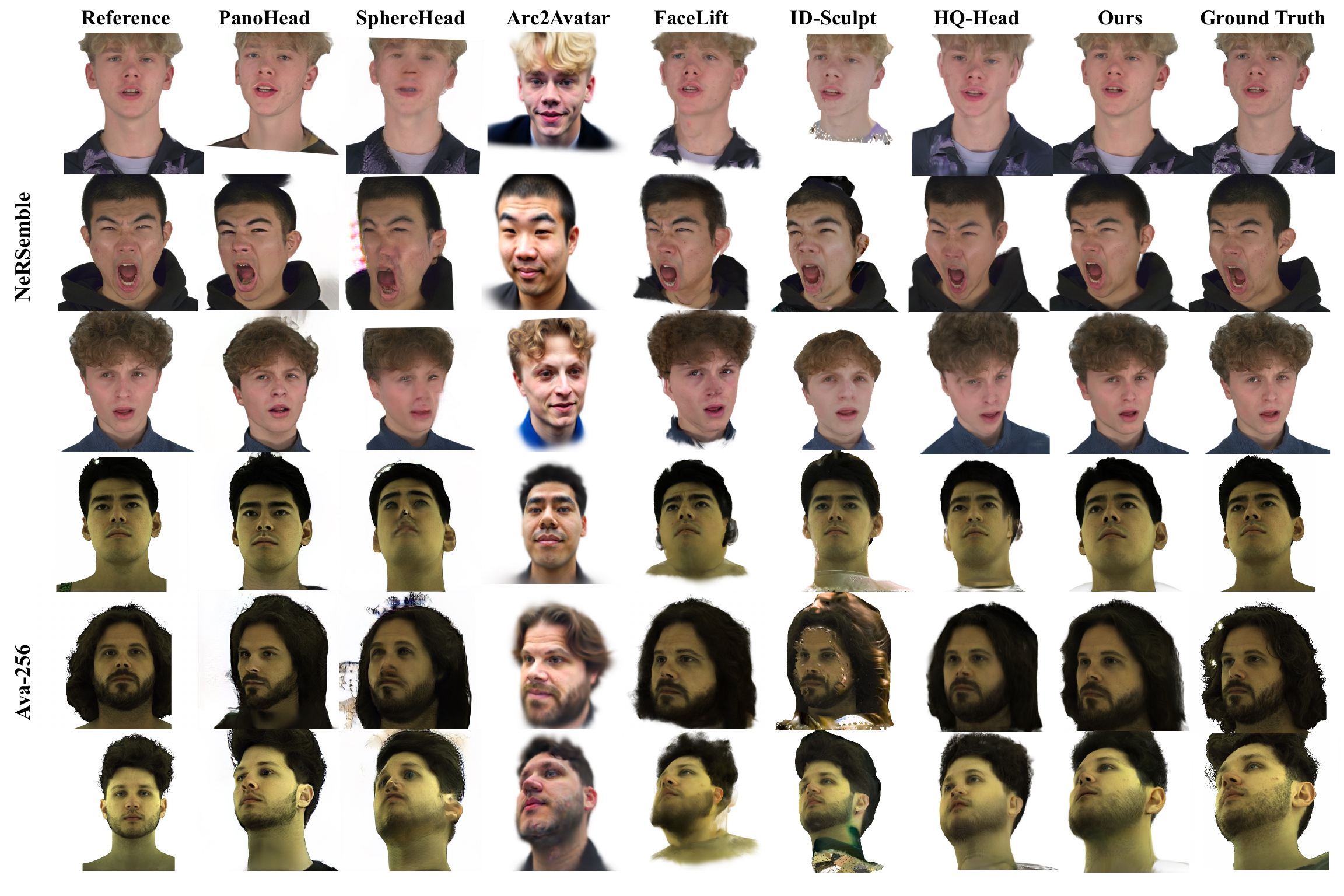}
  \caption{Qualitative visual comparison between Any3DAvatar and other baselines.}
  \label{fig:comparison}
\end{figure*}

\noindent\textbf{Few-Step Denoising:}
In our observation, quality gains saturate after only a few refinements, supporting our hypothesis that the input portrait provides a strong structural prior. We therefore use a small fixed step count $K$. At each step, the current multi-view state is combined with Pl\"ucker embeddings and mapped through the Gaussian representation to predict a clean target:
\begin{equation}
\hat{\mathbf{y}}_{1}^{(k)}=\mathcal{R}\!\left(D_{\mathrm{gs}}\!\left(F_{\theta}(\mathbf{g}_{(t_k)}, t_k, \tilde{\mathbf{x}})\right),\Pi\right),\quad k=0,1,\dots,K-1.
\end{equation}
We then update the state in the image denoising space:
\begin{equation}
\mathbf{y}_{t_{k+1}}=\mathbf{y}_{t_k}+\frac{1}{K}\cdot\frac{\hat{\mathbf{y}}_{1}^{(k)}-\mathbf{y}_{t_k}}{1-t_k},\quad t_k=\frac{k}{K}.
\end{equation}
The last step produces the final Gaussian representation. Empirically, even $K=1$ yields high-quality reconstructions at the lowest inference cost, while a few additional steps further improve fidelity. This flexible schedule provides a practical quality--efficiency trade-off.

\subsection{Auxiliary View-Conditioned Appearance Supervision}
Although strong facial priors enable few-step denoising, latent tokens may still lose high-frequency appearance cues, causing blurred details in views far from the input. We therefore introduce auxiliary view-conditioned appearance supervision (AVAS): alongside Gaussian reconstruction, an auxiliary VAE branch decodes the same DiT tokens into view-conditioned images. This image-space supervision preserves texture and identity cues in weakly observed views and improves novel-view details. The branch is used only during training and removed at inference, incurring no extra inference cost.

\begin{table*}[t]
\centering
\small
\setlength{\tabcolsep}{3pt}
\begin{tabular}{ccccccc|ccccccc}
\hline
\textbf{NeRSemble}   & \textbf{LPIPS}$\downarrow$ & \textbf{PSNR}$\uparrow$ & \textbf{SSIM}$\uparrow$ & \textbf{CSIM}$\uparrow$ & \textbf{DS}$\downarrow$ & \textbf{US}$\uparrow$ & \textbf{Ava-256}     & \textbf{LPIPS}$\downarrow$ & \textbf{PSNR}$\uparrow$ & \textbf{SSIM}$\uparrow$ & \textbf{CSIM}$\uparrow$ & \textbf{DS}$\downarrow$ & \textbf{US}$\uparrow$ \\ \hline
\textbf{PanoHead}    & 0.3241                                      & 13.0231                                  & 0.7446                                   & 0.8128                                   & 0.1274                                   & \underline{4.6874}                     & \textbf{PanoHead}    & 0.3209                                      & 11.3416                                  & 0.7641                                   & \underline{0.8511}                      & 0.1800                                   & 5.2024                                 \\
\textbf{SphereHead}  & 0.3690                                      & 11.6093                                  & 0.7265                                   & 0.6739                                   & 0.2334                                   & 1.5416                                 & \textbf{SphereHead}  & 0.2909                                      & 10.9275                                  & 0.7793                                   & 0.8306                                   & 0.1992                                   & 4.8452                                 \\
\textbf{Arc2Avatar}  & 0.4003                                      & 10.4641                                  & 0.6567                                   & 0.6195                                   & 0.2838                                   & 2.9166                                 & \textbf{Arc2Avatar}  & 0.3488                                      & 9.2206                                   & 0.7176                                   & 0.6990                                   & 0.3785                                   & 2.2292                                 \\
\textbf{FaceLift}    & \underline{0.3056}                          & \underline{13.4421}                      & \underline{0.7558}                       & \underline{0.8237}                       & \underline{0.1107}                       & 3.6458                                 & \textbf{FaceLift}    & \underline{0.2125}                          & \underline{16.1561}                      & \underline{0.8123}                       & 0.8429                                   & \underline{0.0910}                       & \underline{6.2024}                     \\
\textbf{ID-Sculpt}   & 0.3169                                      & 12.0811                                  & 0.7425                                   & 0.7759                                   & 0.1476                                   & 1.7916                                 & \textbf{ID-Sculpt}   & 0.3017                                      & 11.0421                                  & 0.7682                                   & 0.8217                                   & 0.1855                                   & 5.0000                                 \\
\textbf{HQ-Head}     & 0.3661                                      & 12.1271                                  & 0.7270                                   & 0.7636                                   & 0.1311                                   & 3.1250                                 & \textbf{HQ-Head}     & 0.2738                                      & 13.5515                                  & 0.7953                                   & 0.8497                                   & 0.1807                                   & 4.1905                                 \\
\textbf{Any3DAvatar} & \textbf{0.2357}                             & \textbf{16.6202}                         & \textbf{0.7988}                          & \textbf{0.8905}                          & \textbf{0.0621}                          & \textbf{8.1632}                        & \textbf{Any3DAvatar} & \textbf{0.1895}                             & \textbf{18.4251}                         & \textbf{0.8913}                          & \textbf{0.8855}                          & \textbf{0.0694}                          & \textbf{7.8095}                        \\ \hline
\end{tabular}
\caption{Quantitative comparisons on the NeRSemble and Ava-256 datasets.}
\label{tab:comparison_baselines}

\end{table*}

Let $\hat{\mathbf{y}}_{\mathrm{img}}=D_{\mathrm{vae}}(\hat{\mathbf{z}}_g)$ denote the view-conditioned images decoded by the VAE branch, we use a combined pixel and perceptual objective against the ground-truth multi-view images:
\begin{equation}
\mathcal{L}_{\mathrm{AVAS}}=\lambda_{2}\,\mathcal{L}_{2}(\hat{\mathbf{y}}_{\mathrm{img}},\mathbf{y}_{\mathrm{img}})+\lambda_{p}\,\mathcal{L}_{\mathrm{LPIPS}}(\hat{\mathbf{y}}_{\mathrm{img}},\mathbf{y}_{\mathrm{img}}),
\end{equation}
where $\mathcal{L}_{2}$ denotes the $L_2$ reconstruction term and $\mathcal{L}_{\mathrm{LPIPS}}$ denotes the LPIPS perceptual term~\cite{lpips}; $\lambda_{2}$ and $\lambda_{p}$ balance the pixel-level and perceptual terms. The overall training objective is
\begin{equation}
\mathcal{L}_{\mathrm{train}}=\mathcal{L}_{\mathrm{denoise}}+\lambda_{\mathrm{AVAS}}\mathcal{L}_{\mathrm{AVAS}},
\end{equation}
where $\lambda_{\mathrm{AVAS}}$ controls the strength of the auxiliary appearance supervision.

\section{Experiments}
\subsection{Experimental Settings}
\noindent\textbf{Training Setup:} We train the full model on 8 NVIDIA RTX A6000 GPUs for 100 epochs using AdamW optimizer with an initial learning rate of $1\times10^{-4}$ and a cosine annealing schedule. During training, we randomly sample the noise level for each input. The input resolution is set to $H=W=512$ with patch size $p=8$, and the Gaussian color basis uses $\texttt{sh\_degree}=0$. During training, after decoding Gaussian parameters $\hat{\mathbf{g}}_{1}$, we render 8 multi-view images at $512\times512$; together with 4 multi-view images decoded by the VAE branch (also at $512\times512$), they are jointly supervised against ground-truth views. We set $\lambda_{\mathrm{AVAS}}=1$, $\lambda_{2}=1$ and $\lambda_{\mathrm{p}}=0.5$. Unless otherwise specified, we use batch size 6 under the standard training setting. The total training time is approximately 2 days. 

\noindent\textbf{Inference-Time Setup:} For the efficiency comparison, we measure all methods on a single NVIDIA RTX A6000 GPU, excluding one-time model-loading overhead. The reported latency covers the full process from receiving the input image, including preprocessing, to producing the final 3D reconstruction. We use 5 step denoising for inference unless otherwise specified for its best reconstruction quality.

\noindent\textbf{Datasets:} 
We use three datasets for evaluation. For comparison with existing baselines, we select 43 identities with rich expressions from NeRSemble~\cite{nersemble}, using 16 viewpoints per identity to evaluate face reconstruction quality. We also construct an evaluation subset from Ava-256~\cite{ava-256} with 40 identities covering diverse expressions and 80 viewpoints densely surrounding each head. For the ablation studies, we additionally select 40 heads from AnyHead, each with 80 multi-view images, to evaluate the contributions of our model components under comprehensive full-head view coverage.

\begin{figure}
    \centering
    \includegraphics[width=1\linewidth]{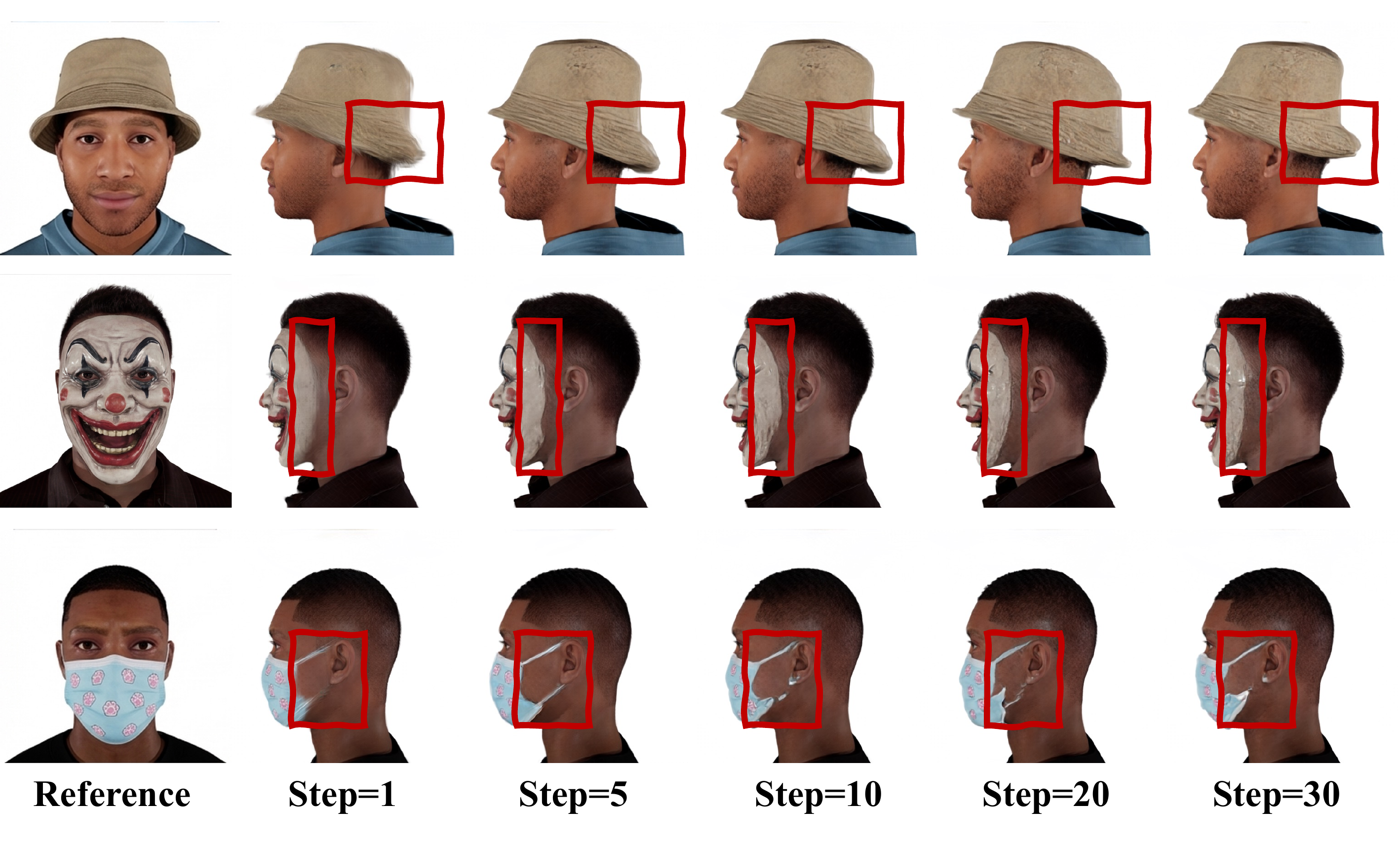}
\caption{Qualitative visualization of the effect of different denoising step counts on the multi-view rendering quality of generated 3D heads.}
    \label{fig:multistep_quality}
\end{figure}

\noindent\textbf{Baselines:} To thoroughly evaluate both reconstruction quality and inference efficiency, we compare our Any3DAvatar with recent single-image full-head reconstruction approaches that provide publicly available code. The selected baselines include PanoHead~\cite{panohead} (CVPR 2023), SphereHead~\cite{spherehead} (ECCV 2024), ID-Sculpt~\cite{id_sculpt} (AAAI 2025), Arc2Avatar~\cite{arc2avatar} (CVPR 2025), FaceLift ~\cite{facelift}(ICCV 2025), and HQ-Head ~\cite{hqhead}(AAAI 2026). For each baseline, we use the best-performing configuration recommended by its official implementation. Since different methods may follow different camera coordinate conventions, we further apply facial-landmark-based alignment after multi-view rendering to ensure consistent view comparison. More comparisons are provided in the supplementary material. 

\noindent\textbf{Evaluation Metric:} Since ground-truth 3D head representations are unavailable, we follow common practice in the field and evaluate reconstruction quality using rendered 2D multi-view images. We evaluate both reconstruction fidelity and identity consistency using five complementary metrics. Specifically, we report LPIPS ~\cite{lpips} and DreamSim(DS) ~\cite{dreamsim} for perceptual similarity, PSNR and SSIM for pixel-level and rendered-view structural fidelity, and cosine similarity of identity embeddings~(CSIM~\cite{arcface}) for identity preservation (higher is better). This combination provides a balanced assessment of visual quality, rendered-view consistency, and identity consistency for multi-view full-head reconstruction. We also conduct an in-depth user study(US) with 32 participants, satisfying common statistical requirements for perceptual evaluation. Participants rate generated examples from different methods on a 0--9 scale for texture quality and identity consistency, yielding 896 ratings in total for comparison.

\subsection{Evaluation against Baselines}
We compare Any3DAvatar(with AVAS) with prior approaches in reconstruction quality and inference efficiency. As shown in Tab.~\ref{tab:times}, our method(denoising step=1) substantially reduces inference time. Quantitative results in Tab.~\ref{tab:comparison_baselines}, together with qualitative comparisons in Fig.~\ref{fig:comparison}, demonstrate the best overall performance on both third-party benchmarks, from the predominantly frontal views of NeRSemble to the dense all-around views of Ava-256, with lower LPIPS/DreamSim and higher CSIM. For more comparisons, please refer to supplementary material and videos.

\begin{table}[]
\centering
\small
\begin{tabular}{cccc} 
\hline
\textbf{Method}   & \textbf{PanoHead }  & \textbf{SphereHead } & \textbf{Arc2Avatar }  \\ 
\hline
\textbf{Time (s) $\downarrow$}    & 130.88            & 127.13             & 8274.67             \\ 
\hline\hline
\textbf{FaceLift } & \textbf{ID-Sculpt } & \textbf{HQ-Head }    & \textbf{Any3DAvatar}        \\ 
\hline
\underline{13.15}            & 3198.79           & 3974.05            & \textbf{0.64}                \\
\hline
\end{tabular}
\caption{Inference time comparison.}
\label{tab:times}
\end{table}

\begin{table}[]
\centering
\small
\setlength{\tabcolsep}{1pt}
\begin{tabular}{ccccccc}
\hline
\textbf{Step}    & \textbf{LPIPS $\downarrow$}              & \textbf{PSNR $\uparrow$}                  & \textbf{SSIM $\uparrow$}                 & \textbf{CSIM $\uparrow$}              & \textbf{DS $\downarrow$}           & \textbf{Time (s) $\downarrow$} \\ \hline
\textbf{step=1}  & 0.2038                          & \underline{18.6689}              & \underline{0.7541}              & 0.8970                          & 0.0690                          & \textbf{0.64} \\ 
\textbf{step=5}  & \textbf{0.2014}                 & \textbf{18.6976}                 & \textbf{0.7545}                 & \textbf{0.8998}                 & \textbf{0.0679}                 & \underline{3.24} \\ 
\textbf{step=10} & \underline{0.2029}              & 18.6000                          & 0.7537                          & \underline{0.8979}              & \underline{0.0684}              & 6.90 \\ 
\textbf{step=20} & 0.2054                          & 18.5606                          & 0.7529                          & 0.8970                          & 0.0697                          & 13.63 \\ 
\textbf{step=30} & 0.2073                          & 18.6142                          & 0.7529                          & 0.8975                          & 0.0705                          & 19.67 \\ \hline
\end{tabular}
\caption{Comparison of different denoising steps on the AnyHead dataset.}
\label{tab:anyhead_multistep}
\end{table}

\subsection{Ablative Analyses}

\noindent\textbf{Effect of Denoising Steps:}
We investigate the effect of denoising steps under the same training (without AVAS) and evaluation settings. As shown in Tab.~\ref{tab:anyhead_multistep} and Fig.~\ref{fig:multistep_quality}, while one-step generation tends to produce over-smoothed, statistically averaged appearances in non-input views, multi-step denoising progressively recovers view-dependent details along a deterministic refinement trajectory, achieving the best overall quality at 5 steps. Further increasing the step count yields no consistent gains and may even degrade generation quality, confirming that the strong structural prior of human heads enables high-quality reconstruction with only a few steps.


\begin{figure}[t]
  \centering
  \includegraphics[width=0.45\textwidth]{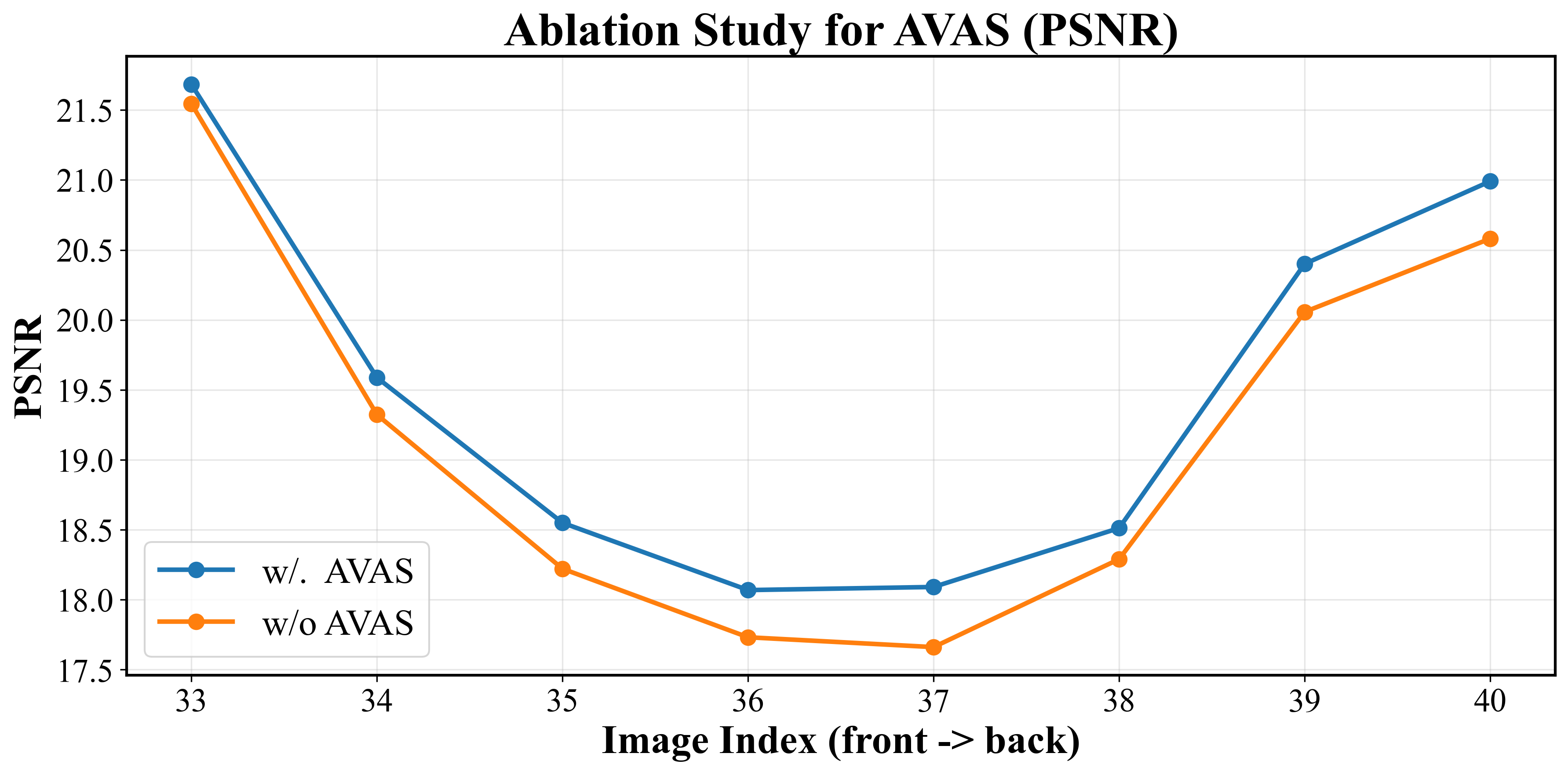}
  \includegraphics[width=0.45\textwidth]{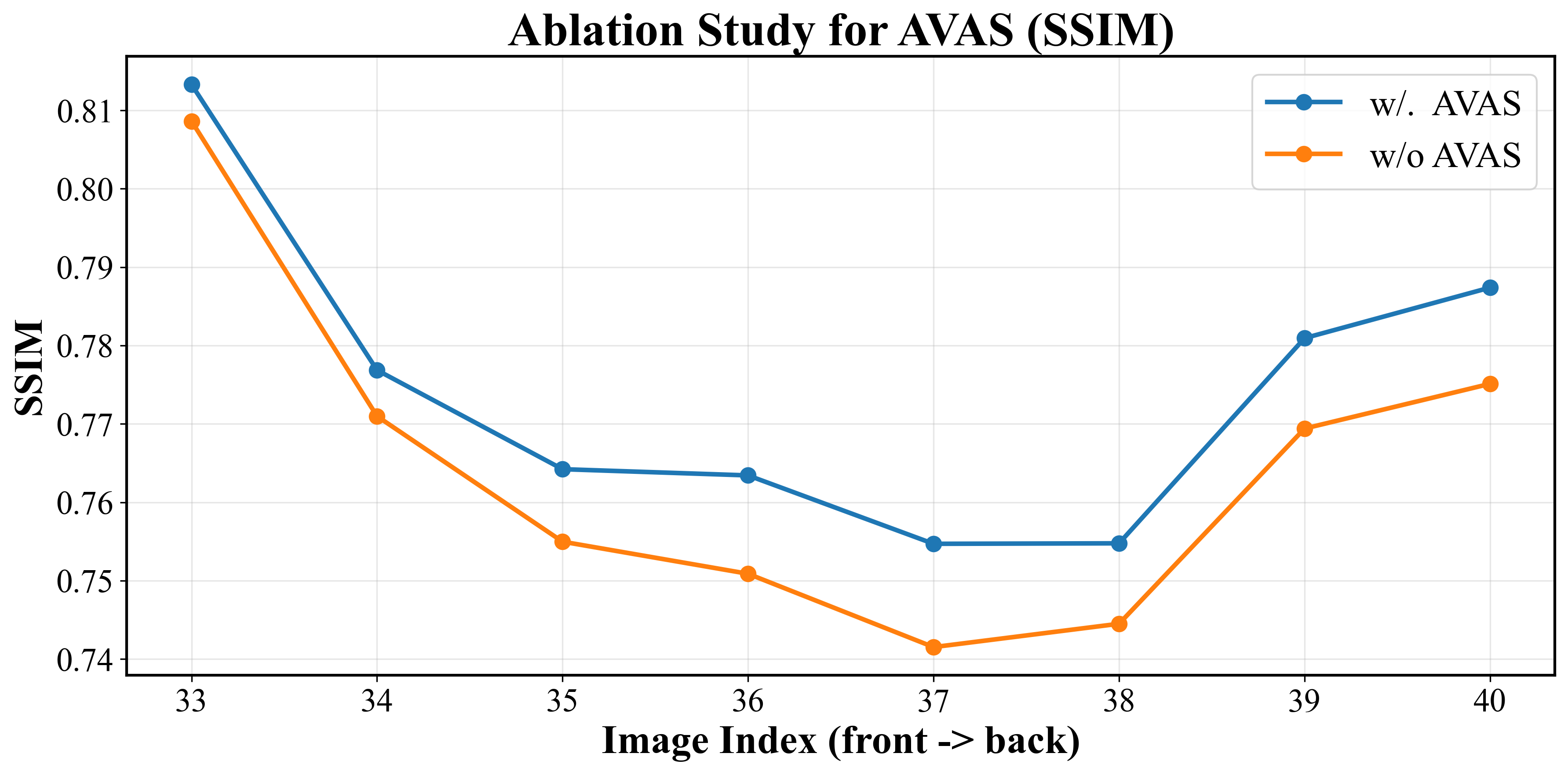}
  \caption{Quantitative comparison of training with and without AVAS on different viewpoint angles (image index).}
  \label{fig:vae_ablation}
\end{figure}

\begin{figure}[t]
  \centering
  \includegraphics[width=0.45\textwidth]{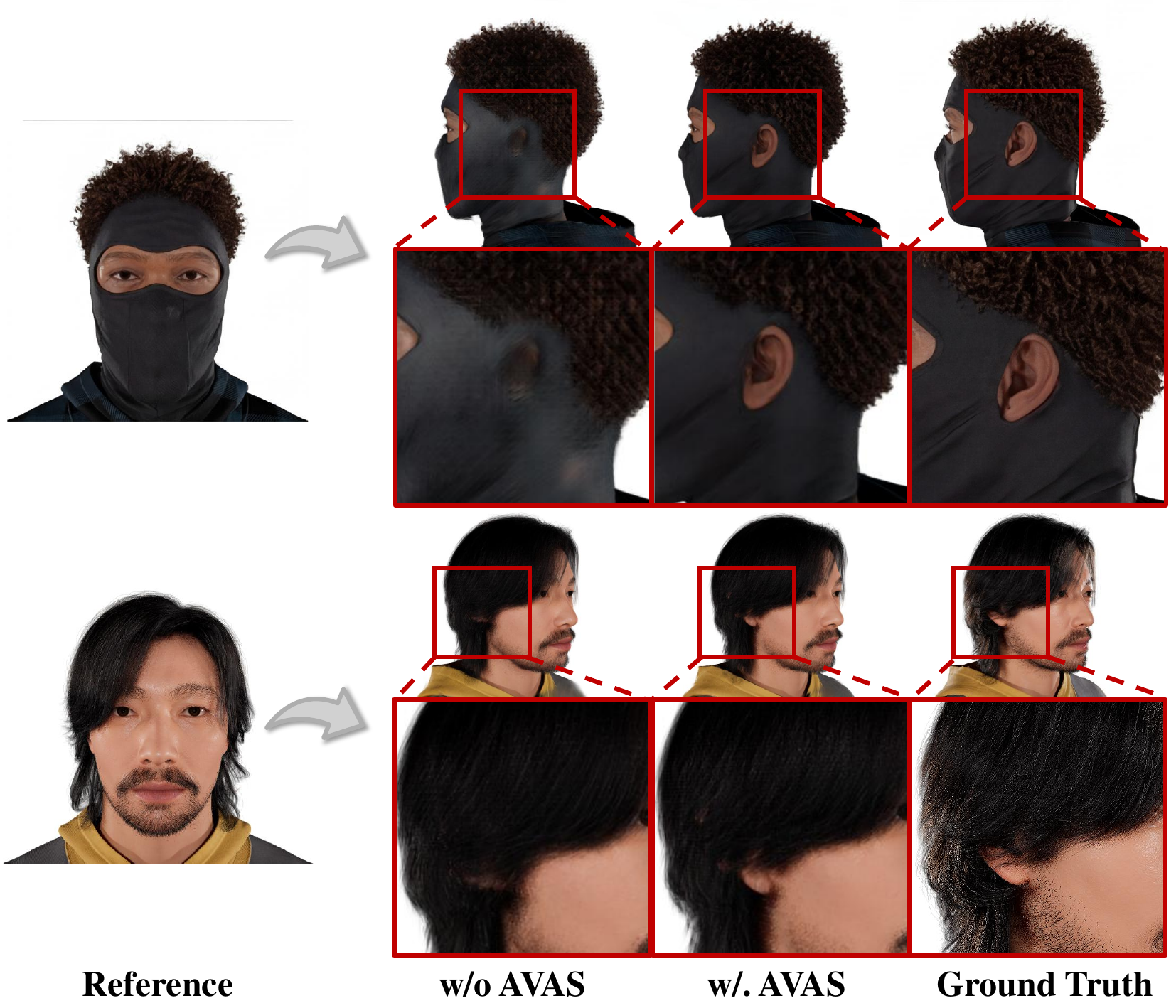}
  \caption{Visual results of AVAS ablation study. In non-input viewpoints, training with AVAS better preserves texture details in non-input views.}
  \label{fig:vae_ablation_visual}
\end{figure}

\begin{figure}[t]
  \centering
  \includegraphics[width=0.5\textwidth]{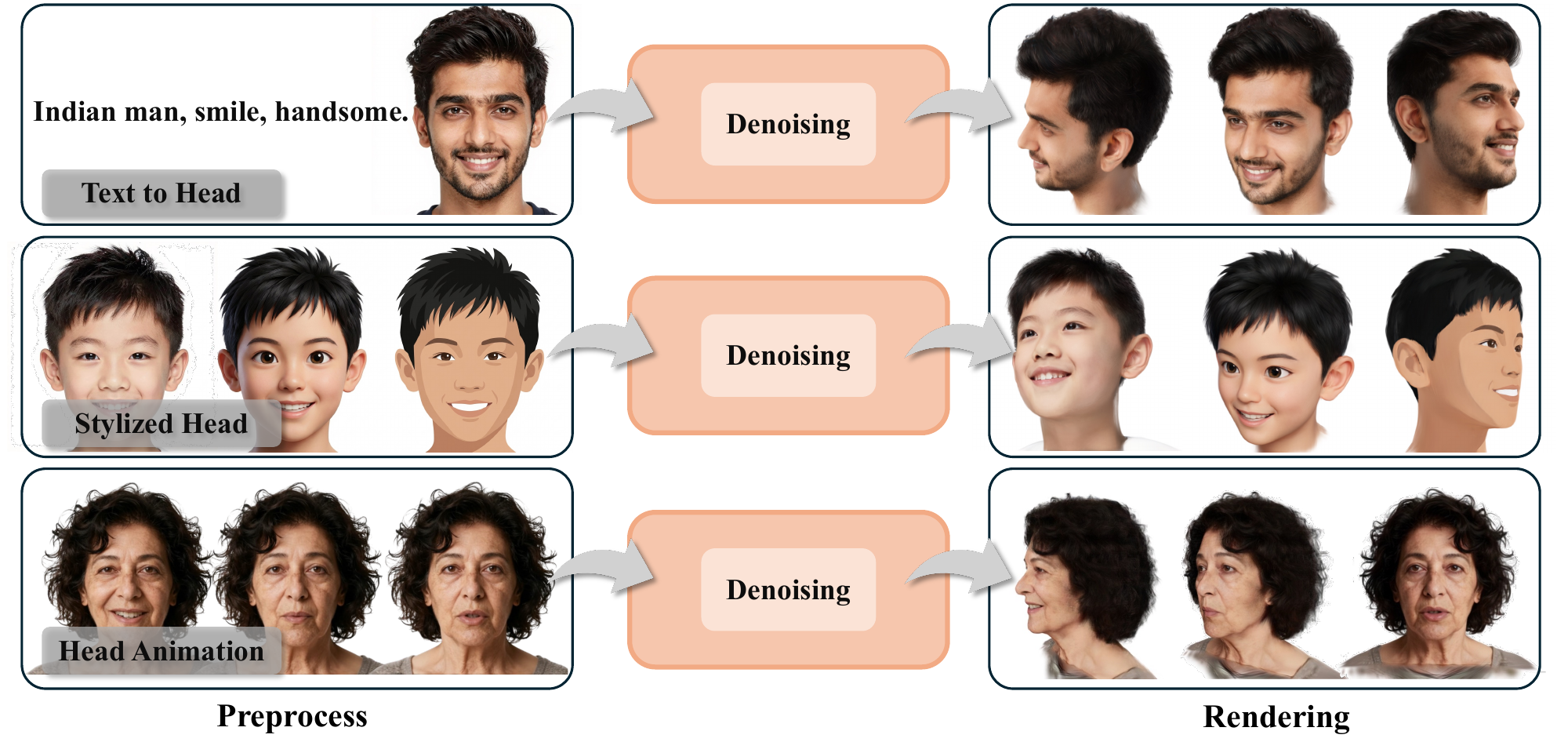}
  \caption{Downstream applications of our method.}
  \label{fig:app}
\end{figure}

\noindent\textbf{Effect of Auxiliary View-Conditioned Appearance Supervision:} We evaluate auxiliary view-conditioned appearance supervision (AVAS) by comparing models trained with and without AVAS, using PSNR and SSIM to measure pixel-level and structural consistency of fine textures. We conduct this study on the AnyHead evaluation subset. As shown in Fig.~\ref{fig:vae_ablation}, adding AVAS consistently improves both metrics, indicating better texture fidelity. The visual examples in Fig.~\ref{fig:vae_ablation_visual} further confirm clearer texture details in novel views.

\noindent\textbf{Comparison without FaceLift-Generated Training Data:}
Part of our training data is constructed using FaceLift, which may raise concerns about the fairness of directly comparing our method with FaceLift. To eliminate this potential advantage, we train a separate variant after removing all FaceLift-generated samples and compare it directly with FaceLift under the same evaluation protocol. As shown in Tab.~\ref{w/o_facelift}, even without FaceLift-generated training data, our method consistently outperforms FaceLift across all metrics on both NeRSemble and Ava-256. These results show that Any3DAvatar's advantage over FaceLift does not rely on training samples generated by FaceLift.

\begin{table}[]
\centering
\small
\setlength{\tabcolsep}{3pt}
\begin{tabular}{cc}
\hline
\textbf{Method}      & \textbf{NeRSemble}                           \\ \hline
\textbf{FaceLift}    & 0.3056/13.4421/0.7558/0.8237/0.1107          \\
\textbf{w/o FaceLift Data} & \textbf{0.2459/16.1976/0.7905/0.8819/0.0659} \\ \hline
\textbf{Method}      & \textbf{Ava-256}                             \\ \hline
\textbf{FaceLift}    & 0.2125/16.1561/0.8123/0.8429/0.0910          \\
\textbf{w/o FaceLift Data} & \textbf{0.1947/18.3987/0.8370/0.8740/0.0810}          \\ \hline
\end{tabular}
\caption{Comparison with FaceLift using our model trained without FaceLift-generated data. Metrics are reported as LPIPS/PSNR/SSIM/CSIM/DreamSim.}
\label{w/o_facelift}
\end{table}

\subsection{Downstream Applications}
Our method supports three applications. (1) Text-to-3D: a text-to-image model first generates a head image that matches a given prompt, which is then fed into our method to reconstruct the corresponding 3D head. (2) Stylized 3D reconstruction: a 2D editing model changes the style of a portrait, and our method reconstructs a stylized 3D head that remains consistent across views, showing its ability to handle different styles. (3) Talking-head video animation: we use the first frame of a talking video to reconstruct a 3D head and render it into a video. A 2D head-driving model~\cite{liveportrait} then uses the talking video as the driving signal to animate the rendered head video, producing the final animated head video. Examples are shown in Fig.~\ref{fig:app}.

\section{Conclusions}
We present Any3DAvatar, an efficient single-image full-head reconstruction framework combining few-step conditional denoising, AVAS for improved novel-view textures without inference overhead, and the diverse, dense-view, accessory-rich AnyHead dataset. Limited multi-view accessory data remains a challenge. Future work will expand such supervision and extend the framework to 4D reconstruction and unconstrained multi-view inputs.

\bibliography{aaai2027}


\appendix

\section*{Appendix}
\section{Overview}
In this supplementary material, we present additional content not covered in the main paper, including:
\begin{itemize}
    \item A pseudocode description of our methods.
    \item Details of our dataset.
    \item Additional experiment results.
    \item Additional visual results.
    \item A discussion of limitations and future work.
\end{itemize}


\begin{algorithm}[t]
\caption{Any3DAvatar Forward Pipeline}
\label{alg:any3davatar}
\begin{algorithmic}[1]
\REQUIRE Identity $\mathcal{I}$, portrait $\mathbf{x}\in\mathbb{R}^{3\times H\times W}$, patch size $p$ with $p\mid H$ and $p\mid W$
\ENSURE Predicted four-view images $\mathbf{y}_{\mathrm{img}}$, Gaussian parameters $\hat{\mathbf{g}}_{1}$, rendered Gaussian views $\mathbf{y}_{\mathrm{GS}}$
\STATE Compute Pl\"ucker rays $\mathbf{r}\in\mathbb{R}^{6\times H\times W}$ for $\mathbf{x}$ and build $\tilde{\mathbf{x}}=[\mathbf{x};\mathbf{r}]\in\mathbb{R}^{9\times H\times W}$
\STATE Tokenize and encode context features $\mathbf{ctx}=E_{\mathrm{img}}(\tilde{\mathbf{x}})\in\mathbb{R}^{N_h\times N_w\times 1024}$, where $N_h=\lfloor H/p\rfloor$, $N_w=\lfloor W/p\rfloor$
\STATE Set canonical view set $\mathcal{V}=\{\mathrm{front},\mathrm{left},\mathrm{right},\mathrm{back}\}$ and view rays $\mathbf{R}\in\mathbb{R}^{4\times H\times W\times 6}$
\STATE Set diffusion timestep $t$:
\FORALL{$v\in\mathcal{V}$}
\STATE Given clean multi-
view images $y_1^{(v)}$ and image noise $y_0^{(v)}$, $y_t^{(v)}=(1-t)y_0^{(v)}+ty_1^{(v)}$  and concatenate with the corresponding Pl\"ucker rays to form $\mathbf{g}_{t}^{(v)}\in\mathbb{R}^{H\times W\times 9}$
\STATE Encode view-wise Gaussian tokens $\mathbf{z}_{g}^{(v)}=E_{\mathrm{GS}}(\mathbf{g}_{t}^{(v)})\in\mathbb{R}^{N_h\times N_w\times 1024}$
\ENDFOR
\STATE Stack all views as $\mathbf{z}_{g}\in\mathbb{R}^{4\times N_h\times N_w\times 1024}$
\STATE Flatten both branches: $\hat{\mathbf{c}}=\mathrm{Flatten}(\mathbf{ctx})\in\mathbb{R}^{(N_hN_w)\times 1024}$, $\hat{\mathbf{g}}=\mathrm{Flatten}(\mathbf{z}_{g})\in\mathbb{R}^{(4N_hN_w)\times 1024}$
\STATE Concatenate tokens along sequence dimension: $\mathbf{s}_t=[\hat{\mathbf{c}};\hat{\mathbf{g}}]\in\mathbb{R}^{(5N_hN_w)\times 1024}$

\STATE Process sequence with DiT: $\mathbf{s}=F_{\mathrm{DiT}}(\mathbf{s}_t,t)$
\STATE Split denoised Gaussian tokens $\tilde{\mathbf{z}}_{g}$ from $\mathbf{s}$
\STATE Reshape $\tilde{\mathbf{z}}_{g}\rightarrow\hat{\mathbf{z}}_{g}$ and decode four-view RGB predictions: $\mathbf{y}_{\mathrm{img}}=D_{\mathrm{vae}}(\hat{\mathbf{z}}_{g})\in\mathbb{R}^{4\times H\times W\times 3}$ \COMMENT{training only}
\STATE Decode Gaussian parameters: $\hat{\mathbf{g}}_{1}=D_{\mathrm{gs}}(\tilde{\mathbf{z}}_{g})\in\mathbb{R}^{(4HW)\times 14}$
\STATE Assemble Gaussian point cloud from $\hat{\mathbf{g}}_{1}$ and render multi-view images $\hat{\mathbf{y}}_{1}$
\STATE Inference: skip $D_{\mathrm{vae}}(\cdot)$ and use only $D_{\mathrm{gs}}(\cdot)$ for decoding
\end{algorithmic}
\end{algorithm}

\section{Pseudocode}

Following the notation defined in Section 3.1 of the main paper, we present the pseudocode of our method in Algorithm ~\ref{alg:any3davatar}.
The VAE decoding branch is used only during training for auxiliary image-space supervision, while inference uses only the Gaussian decoding branch.

\subsection{Model Architecture and Configuration}
Our DiT backbone consists of 24 transformer layers, with a hidden dimension of 1024 and 64 attention heads in each layer. The input portrait and the Gaussian denoising inputs are partitioned into non-overlapping patches with a patch size of $p=8$. At the input resolution of $512\times512$, this patch embedding produces a $64\times64$ token grid for each view.

\subsection{Pl\"ucker Ray Construction and Noise Concatenation}
For each pixel $(u,v)$, we first compute its corresponding camera ray in world coordinates. Let $\tilde{\mathbf{p}}_{u,v}=[u,v,1]^{\top}$, let $\mathbf{K}$ denote the camera intrinsic matrix, and let $[\mathbf{R}\mid\mathbf{t}]$ denote the world-to-camera extrinsics, such that $\mathbf{x}_{c}=\mathbf{R}\mathbf{x}_{w}+\mathbf{t}$. The camera center $\mathbf{o}$ and the normalized ray direction $\mathbf{d}_{u,v}$ are
\begin{equation}
\mathbf{o}=-\mathbf{R}^{\top}\mathbf{t}, \qquad
\mathbf{d}_{u,v}=\frac{\mathbf{R}^{\top}\mathbf{K}^{-1}\tilde{\mathbf{p}}_{u,v}}
{\left\|\mathbf{R}^{\top}\mathbf{K}^{-1}\tilde{\mathbf{p}}_{u,v}\right\|_{2}}.
\end{equation}
We represent this oriented ray using its six-dimensional Pl\"ucker coordinate
\begin{equation}
\mathbf{r}_{u,v}=\left(\mathbf{d}_{u,v},\mathbf{m}_{u,v}\right), \qquad
\mathbf{m}_{u,v}=\mathbf{o}\times\mathbf{d}_{u,v},
\end{equation}
where $\mathbf{m}_{u,v}$ is the moment of the ray. Computing these coordinates for all pixels gives a Pl\"ucker ray map $\mathbf{R}^{(v)}\in\mathbb{R}^{H\times W\times6}$ for each canonical view $v$.

During training, we construct the three-channel noisy image state for view $v$ as $\mathbf{y}_{t}^{(v)}=(1-t)\mathbf{y}_{0}^{(v)}+t\mathbf{y}_{1}^{(v)}$, where $\mathbf{y}_{0}^{(v)}$ is image-space Gaussian noise and $\mathbf{y}_{1}^{(v)}$ is the clean target view. We then concatenate the noisy image state and its Pl\"ucker ray map along the channel dimension:
\begin{equation}
\mathbf{g}_{t}^{(v)}=\mathrm{Concat}\!\left(\mathbf{y}_{t}^{(v)},\mathbf{R}^{(v)}\right)
\in\mathbb{R}^{H\times W\times9}.
\end{equation}
At $t=0$, the image component is pure Gaussian noise, while the six ray channels remain fixed across timesteps. We analogously concatenate the input portrait with its ray map to form $\tilde{\mathbf{x}}=[\mathbf{x};\mathbf{r}]\in\mathbb{R}^{9\times H\times W}$. This construction associates every image or noise pixel with an oriented ray in 3D space, providing the Gaussian tokenizer with an explicit spatial prior.

\begin{figure}[t]
    \centering
    \includegraphics[width=\linewidth]{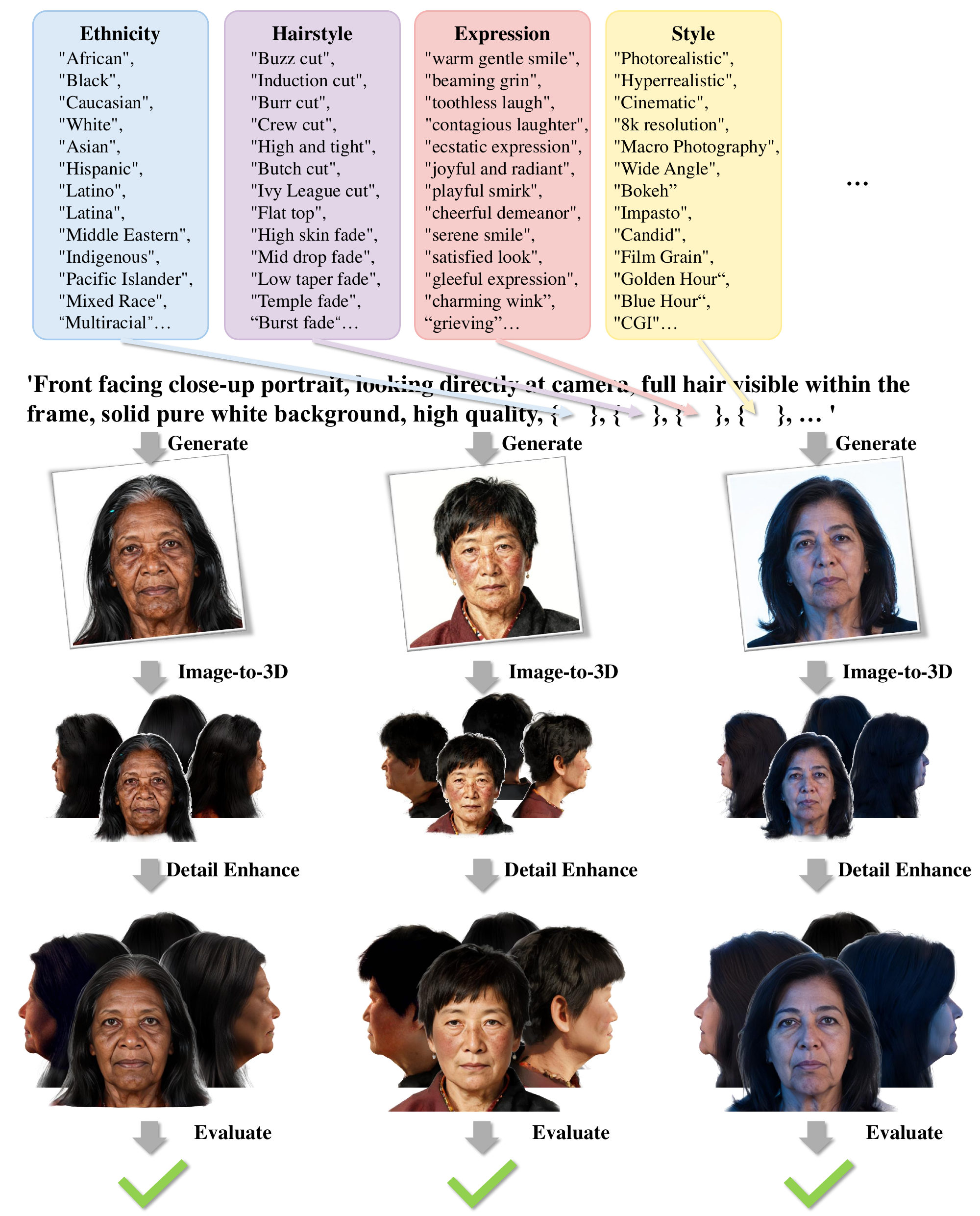}
    \caption{Examples of AI-generated heads used in our dataset construction pipeline.}
    \label{fig:ai_generated_heads}
\end{figure}

\section{Dataset Details}

In this section, we provide a detailed description of AnyHead dataset construction pipeline, including data collection, filtering, and quality control. We also present a comparative analysis against existing datasets, highlighting the main strengths of our dataset in identity diversity, view coverage, and fine-grained appearance fidelity.

\subsection{Construction Details}
\noindent\textbf{AI-Generated Heads:}
To reduce the selection bias and identity-coverage limitations of real-captured multi-view head datasets, we build an AI-generated subset with nearly 1,000 identities and diverse attributes. As shown in Fig.~\ref{fig:ai_generated_heads}, the pipeline contains five steps: (i) building an attribute lexicon for each identity-related factor (e.g., age, ethnicity, hairstyle, expression, and lighting) and uniformly sampling from these lexicons, (ii) generating frontal portraits using a text-to-image model~\cite{z-image} under a fixed-view setting, where prompts are composed of a shared background template and personalized attribute prompts to guide high-quality and diverse head generation, (iii) converting the generated portraits into 3D Gaussian heads with an image-to-3D model~\cite{facelift} and rendering 80 multi-view images around the head with elevation angles from $-20^\circ$ to $20^\circ$, (iv) enhancing facial details via an image restoration model~\cite{codeformer} to suppress Gaussian artifacts and improve facial details, and (v) removing low-quality samples using automatic realism checks driven by a multimodal large language model~\cite{qwen3vl}. All rendered multi-view images in this subset have a resolution of $1024\times1024$. This process improves identity diversity and reduces sampling bias for robust training.

\begin{figure}[t]
    \centering
    \includegraphics[width=\linewidth]{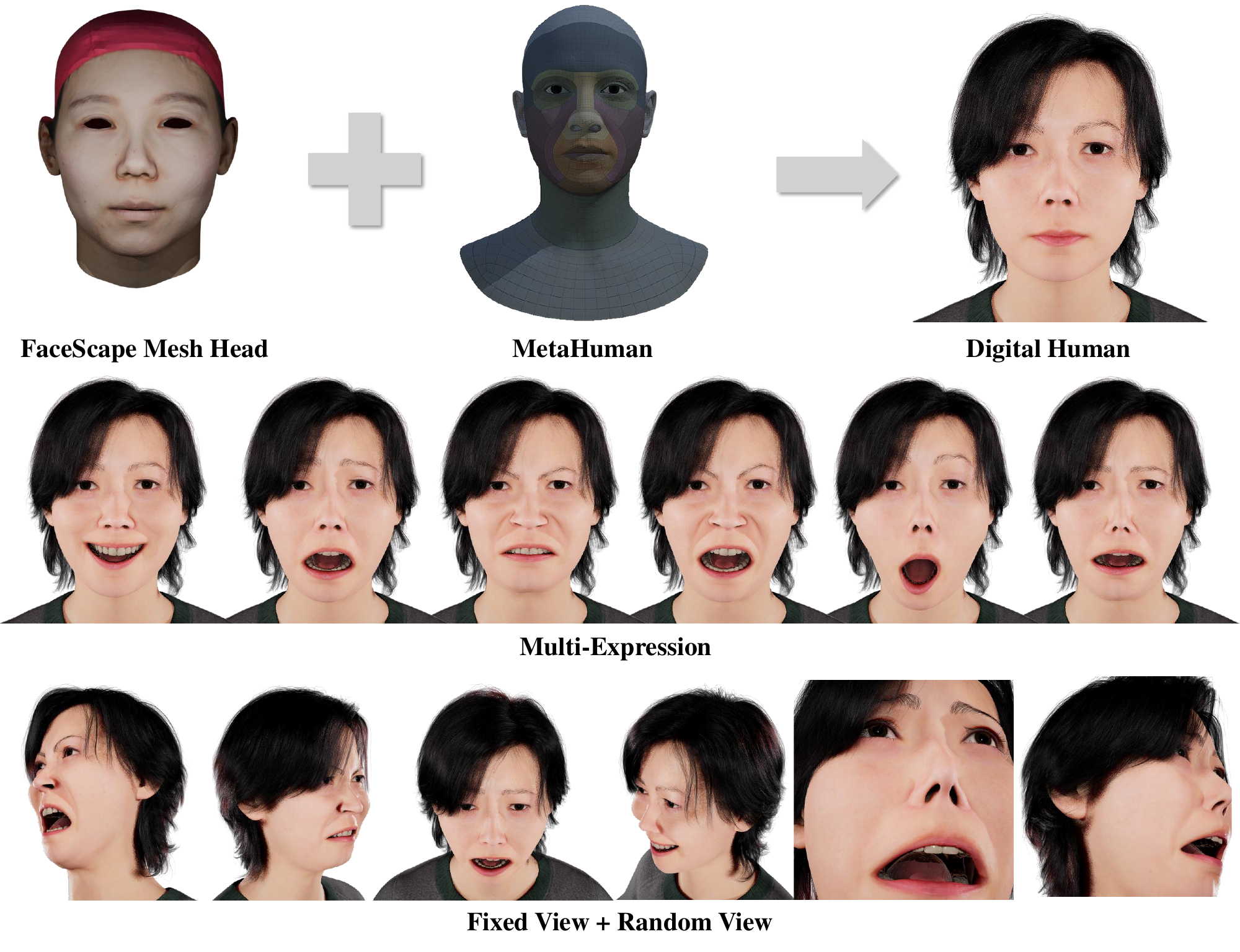}
    \caption{Examples of digital human heads rendered in UE MetaHuman with diverse appearance attributes and expressions.}
    \label{fig:digital_human_heads}
\end{figure}

\noindent\textbf{Digital Human Heads:}
While AI-generated data improves identity diversity, real-captured datasets still provide limited viewpoint coverage due to constrained camera quantity and placement. To address this limitation, we use Unreal Engine (UE) to synthesize high-fidelity digital human heads with dense, controllable camera sampling (Fig.~\ref{fig:digital_human_heads}). We first sample real head keypoints from FaceScape~\cite{facescape}, and then construct multi-expression digital heads conditioned on these keypoints. The FaceScape dataset was captured from real participants, with authorization obtained for the use of their portrait data.
Specifically, for each identity, we fit and retarget facial geometry in UE MetaHuman according to sampled keypoints, and then assign appearance attributes to the MetaHuman mesh, including skin tone, hairstyle, eyebrow/beard style, and material-level facial details, to obtain high-fidelity digital heads with realistic texture and identity-consistent appearance. 
To model expression diversity, we further drive the digital heads with the UE MetaHuman facial rig/skeleton system, enabling controllable synthesis of different expressions for the same identity.
We adopt two complementary view-sampling strategies: (i) \emph{fixed circular views}, where 40 cameras are placed at equal distance around the head across elevation angles from $-20^\circ$ to $20^\circ$ to provide stable global geometric supervision, and (ii) \emph{random views}, where 40 cameras are sampled with random positions and orientations to emphasize local regions and fine-grained details (e.g., hair boundary, ears, and profile contours). All rendered multi-view images in this subset have a resolution of $1024\times1024$.

\begin{figure}[t]
    \centering
    \includegraphics[width=\linewidth]{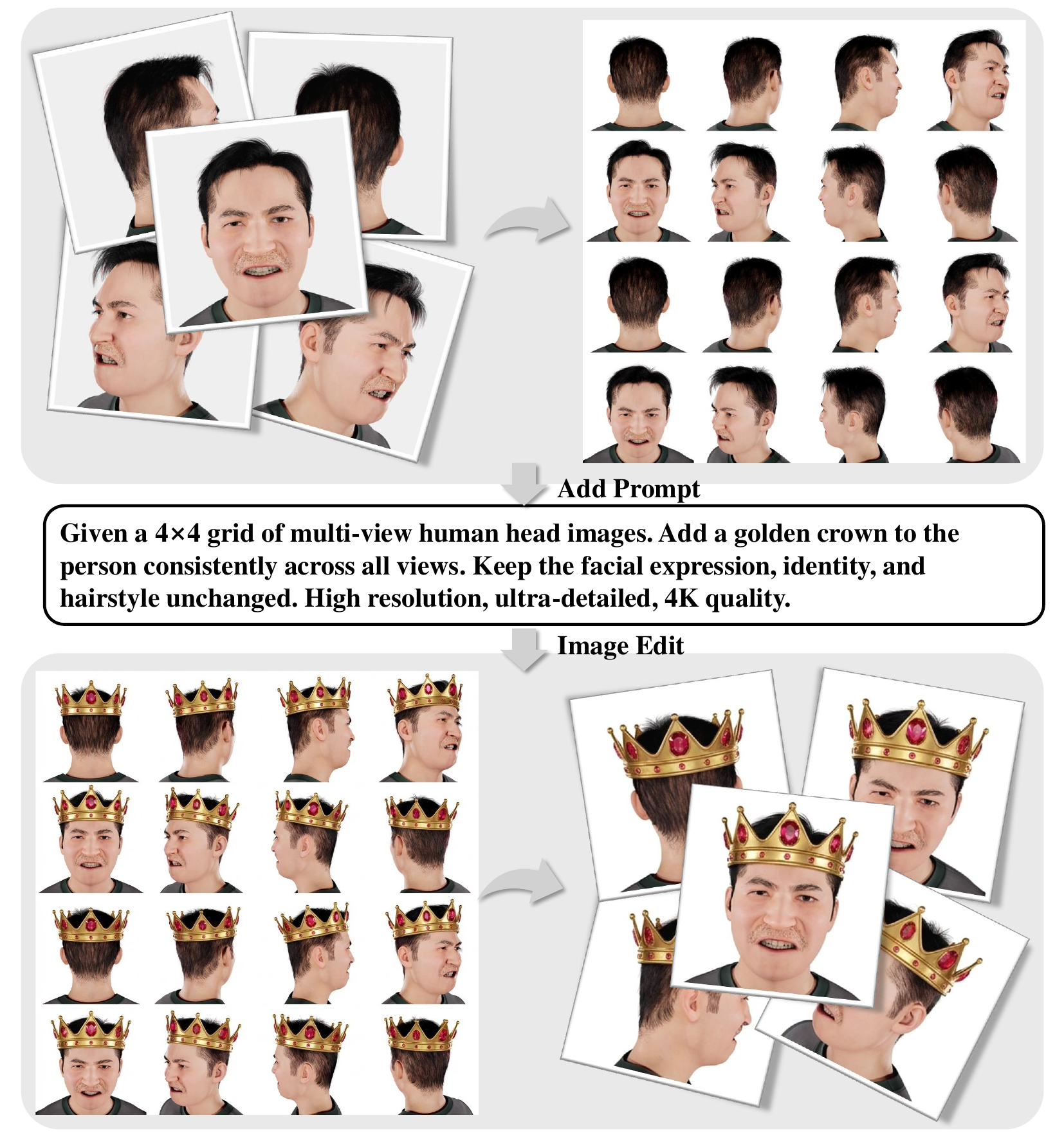}
    \caption{Examples of accessory-rich heads generated by controlled multi-view accessory augmentation.}
    \label{fig:accessory_heads}
\end{figure}

\noindent\textbf{Accessory-Rich Heads:}
Existing head datasets often underrepresent accessories. To bridge this gap, we construct an accessory-rich multi-view subset through controlled accessory augmentation of Digital Human Heads (Fig.~\ref{fig:accessory_heads}). Specifically, for each identity, we first arrange sampled multi-view renders into a single multi-view grid so that a single edit can be enforced jointly across views. The selected views include 8 views at horizontal elevation ($0^\circ$), plus 4 views at $-10^\circ$ and 4 views at $10^\circ$.
We then edit the grid with Nano Banana Pro to add accessories (e.g., glasses, hats, and other wearable items), which provides strong cross-view consistency while preserving facial identity and head geometry. During editing, we use a \emph{fixed prompt + random accessory description} strategy: the fixed prompt explicitly embeds multi-view consistency constraints (same identity, same geometry, and consistent accessory shape/position across views), while random accessory descriptions are injected to increase accessory diversity.
After editing, we perform manual quality control to remove samples with noticeable cross-view inconsistency in identity, geometry, or accessory appearance (e.g., missing frames, drifting accessory position, or shape mismatch across views). Finally, each qualified edited grid is split back into individual views to recover high-quality multi-view tuples with accessory annotations inherited from the edited grid. All resulting multi-view images in this subset have a resolution of $1024\times1024$.

\subsection{Data Organization}

The dataset is organized into three subsets with aligned but task-specific directory structures. \textbf{AI\_Generated\_Heads\_Dataset} stores per-identity multi-view RGB images under each \{HEAD\_ID\} folder, where \{VIEW\_ID\}.png denotes one rendered viewpoint and README.md documents the split and metadata format. Digital human heads adds an additional \{EXPRESSION\_ID\} level under each identity to separate different facial expressions, and provides camera parameters file for each identity/expression group to record camera calibration and view definitions. Accessory-Rich heads follows a similar hierarchy but replaces the expression level with \{ACCESSORY\_ID\}, so that each accessory condition is grouped independently while preserving multi-view images and associated camera parameters.

\dirtree{%
.1 AI\_Generated\_Heads\_Dataset/.
.2 \{HEAD\_ID\}/.
.3 \{VIEW\_ID\}.png.
.3 ....
.2 \{HEAD\_ID\}/.
.3 \{VIEW\_ID\}.png.
}

\dirtree{%
.1 Digital\_Human\_Heads\_Dataset/.
.2 \{HEAD\_ID\}/.
.3 \{EXPRESSION\_ID\}/.
.4 \{VIEW\_ID\}.png.
.4 ....
.3 CAMERA\_PARAMETERS.json /.
.2 \{HEAD\_ID\}/.
.3 \{EXPRESSION\_ID\}/.
.4 \{VIEW\_ID\}.png.
.4 ....
.3 CAMERA\_PARAMETERS.json /.
}

\dirtree{%
.1 Accessory\_Rich\_Heads\_Dataset/.
.2 \{HEAD\_ID\}/.
.3 \{ACCESSORY\_ID\}/.
.4 \{VIEW\_ID\}.png.
.4 ....
.3 CAMERA\_PARAMETERS.json /.
.2 \{HEAD\_ID\}/.
.3 \{ACCESSORY\_ID\}/.
.4 \{VIEW\_ID\}.png.
.4 ....
.3 CAMERA\_PARAMETERS.json /.
}

\subsection{Comparison with Other Datasets}

We compare AnyHead against three representative head datasets, including Ava-256~\cite{ava-256}, NeRSemble~\cite{nersemble}, and RenderMe-360~\cite{renderme360}. To focus on dataset richness, we evaluate the compared datasets using the following indicators: (i) identity number, (ii) accessory support, (iii) full-head coverage, (iv) camera-view count, (v) stylized-head support, and (vi) detailed-view availability.

\begin{table*}[t]
\centering

\begin{tabular}{lcccccc}
\toprule
\textbf{Dataset} & \textbf{ID Number} & \textbf{Accessory} & \textbf{Full Head} & \textbf{Camera View} & \textbf{Stylized Head} & \textbf{Detailed View} \\
\midrule
\textbf{Ava-256} & 256 & \xmark & \cmark & 80 & \xmark & \cmark \\
\textbf{NeRSemble} & 418 & \xmark & \xmark & 16 & \xmark & \xmark \\
\textbf{RenderMe-360} & 500 & \cmark & \cmark & 60 & \xmark & \xmark \\
\textbf{AnyHead} & 1060 & \cmark & \cmark & 80 & \cmark & \cmark \\
\bottomrule
\end{tabular}
\caption{Comparison with existing head datasets.}
\label{tab:dataset_comparison}
\end{table*}


\begin{table*}[]

\centering
\small
\setlength{\tabcolsep}{2pt}
\begin{tabular}{cccccccc}
\hline
\textbf{Method}                        & \textbf{PanoHead} & \textbf{SphereHead} & \textbf{Arc2Avatar}    & \textbf{FaceLift}       & \textbf{ID-Sculpt}      & \textbf{HQ-Head}                & \textbf{Any3DAvatar}   \\ \hline
\textbf{Render Time (s)$\downarrow$}                     & 0.5628            & \underline{0.7532}              & $2.0444\times 10^{-3}$ & $3.8667 \times 10^{-3}$ & $3.0602 \times 10^{-3}$ & $\mathbf{1.8993\times 10^{-3}}$ & $4.4328\times 10^{-3}$ \\
\textbf{Infer + Render Time (s)$\downarrow$} & 131.4428          & 127.8832            & \underline{8274.6720}              & 13.1539                 & 3198.7931               & 3974.0519                       & \textbf{0.6544}                 \\
\textbf{File Type}                     & implicit          & implicit            & gs                     & gs                      & mesh                    & gs                              & gs                     \\ \hline
\end{tabular}
\caption{Comparison of runtime with full-head reconstruction methods.}
\label{tab:render_efficiency}
\end{table*}

\begin{table}[]
\centering
\small
\setlength{\tabcolsep}{2pt}
\begin{tabular}{cccccc}
\hline
\textbf{Method} & \textbf{LPIPS} $\downarrow$ & \textbf{PSNR} $\uparrow$ & \textbf{SSIM} $\uparrow$ & \textbf{CSIM} $\uparrow$ & \textbf{DS} $\downarrow$ \\ \hline
\textbf{TRELLIS} & \underline{0.3254} & \underline{11.7264} & 0.6539 & \underline{0.7643} & \underline{0.2281} \\
\textbf{Open-Diffusion-GS} & 0.3191 & 12.2962 & \underline{0.6443} & 0.7660 & 0.1680 \\
\textbf{Any3DAvatar} & \textbf{0.1894} & \textbf{19.8639} & \textbf{0.7632} & \textbf{0.9258} & \textbf{0.0633} \\ \hline
\end{tabular}
\caption{Quantitative comparison with image-to-3D methods on the AnyHead dataset.}
\label{tab:image_to_3d_comparison}
\end{table}

As shown in Table~\ref{tab:dataset_comparison}, AnyHead achieves the best overall richness. It contains the largest identity set (1060), maintains dense multi-view coverage (80 views), and is the only dataset that simultaneously supports accessories, full-head capture, stylized heads, and detailed views. In contrast, existing datasets usually satisfy only part of these requirements. This demonstrates that our dataset provides the most comprehensive data richness for training and evaluating robust full-head reconstruction models.

\subsection{Intended Usage and Licensing}

This dataset is constructed for generating and editing high-quality 3D human heads, and is intended to support related research on robust reconstruction, rendering, and controllable avatar creation. We plan to release the dataset publicly. The release will follow the Apache License 2.0.

\begin{figure}[t]
    \centering
    \includegraphics[width=\linewidth]{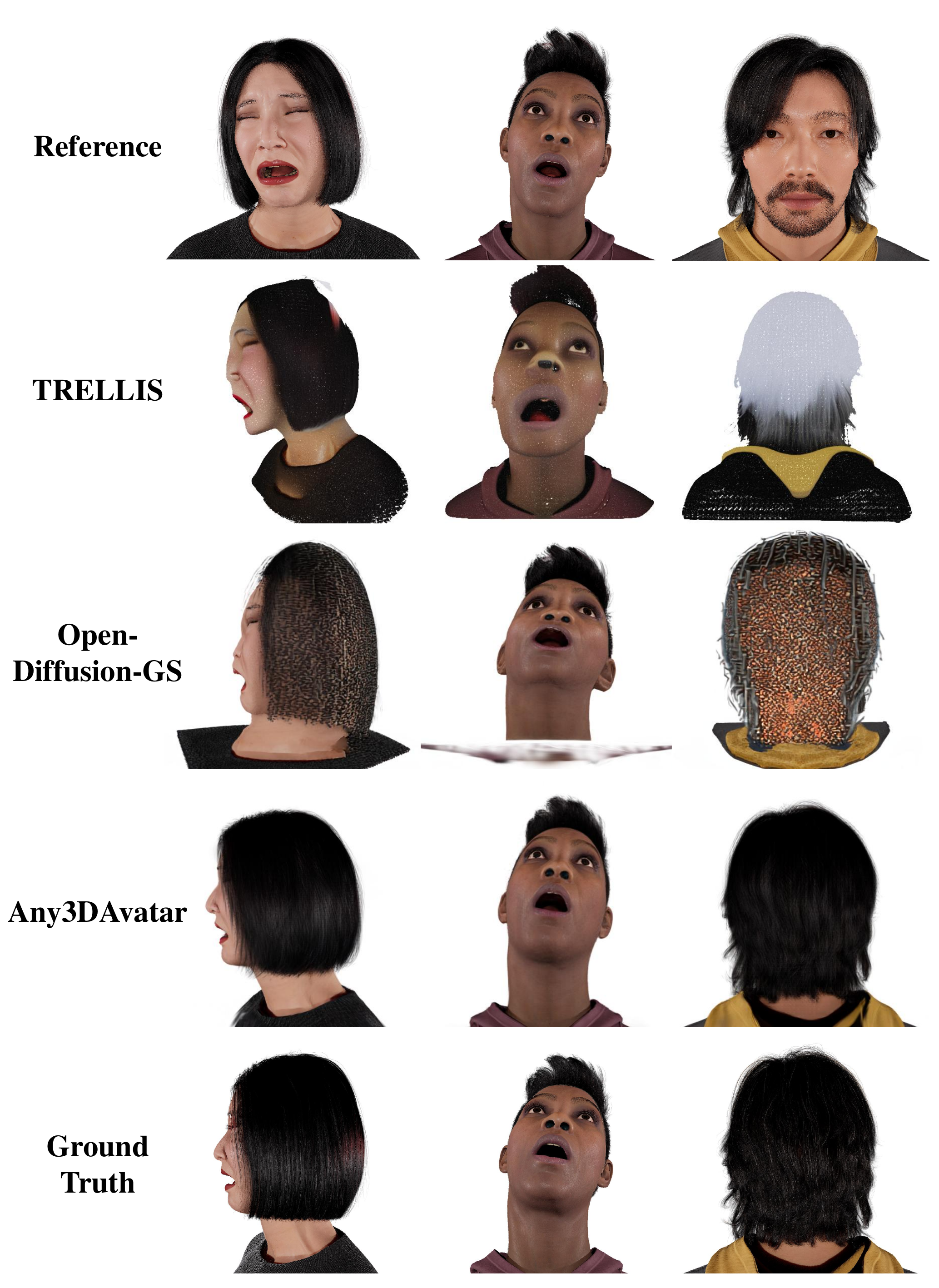}
    \caption{Visual comparison with image-to-3D methods.}
    \label{fig:comparison_123d}
\end{figure}

\section{Additional Experiments}

In this section, we provide additional experimental studies to further validate our method. Specifically, we include rendering efficiency evaluation, comparisons with general image-to-3D Gaussian point cloud models, comparisons with other single-image 3D head reconstruction methods, and more ablation studies on key design choices. Unless otherwise stated, all experimental settings, metrics, and implementation details are kept the same as those in the main paper for fair comparison.

\subsection{Rendering Efficiency Comparison}

From Table~\ref{tab:render_efficiency}, we observe a clear efficiency trend across representation types. All runtime measurements are conducted at a resolution of $512\times512$ on a server equipped with an NVIDIA A6000 GPU. Methods based on implicit representations generally show the slowest rendering speed, while methods based on explicit Gaussian splats or mesh representations render substantially faster. In particular, Any3DAvatar achieves very fast per-frame rendering, remaining competitive with other explicit-representation methods. More importantly, when considering the total runtime of inference plus rendering, our method is the fastest among all compared methods by a large margin. These results highlight that Any3DAvatar delivers the best overall efficiency for practical single-image full-head reconstruction and visualization.

\subsection{AVAS Details}
In the main paper, we show that AVAS improves detail consistency in regions unobserved from the input view and report view-wise PSNR and SSIM to analyze its effect across viewpoints. Here, we further compare models trained with and without AVAS using the complete set of evaluation metrics, including LPIPS, PSNR, SSIM, CSIM, and DreamSim. As 
shown in Tab. \ref{tab:avas}, AVAS consistently improves performance across all metrics, demonstrating its effectiveness in enhancing the quality of full-head reconstruction.

\subsection{User Study Details}
We provide additional details of the user study in the main paper. We conducted a double-blind user study with 32 randomly recruited participants. The method labels are anonymous, and both test cases and the 7 methods were randomly ordered to avoid bias. Across 896 ratings, the Friedman test shows significant differences ($\chi^2(6)=84.15$, $p<0.001$), and ICC(2,k)=0.977 indicates highly consistent and reliable preferences.

\begin{table}[]
\centering
\small
\setlength{\tabcolsep}{2pt}
\begin{tabular}{cccccc}
\hline
\textbf{Strategy} & \textbf{LPIPS}$\downarrow$  & \textbf{PSNR}$\uparrow$    & \textbf{SSIM}$\uparrow$   & \textbf{CSIM}$\uparrow$   & \textbf{DS} $\downarrow$    \\ \hline
\textbf{w/o AVAS} & 0.2014          & 18.6976          & 0.7545          & 0.8998          & 0.0679          \\
\textbf{w/ AVAS}  & \textbf{0.1894} & \textbf{19.8639} & \textbf{0.7632} & \textbf{0.9258} & \textbf{0.0633} \\ \hline
\end{tabular}
\caption{Ablation study on the effectiveness of the proposed AVAS strategy.}
\label{tab:avas}
\end{table}

\subsection{Multi-View 2D IoU Metric}
We additionally evaluate geometric reconstruction quality using multi-view 2D Intersection over Union (IoU). Specifically, we render binary foreground masks of the reconstructed and reference head geometry from the same set of camera viewpoints. We then compute the IoU between the two silhouette masks for each view and report the average across all evaluated views. Although the IoU is computed in the 2D image plane, aggregating it over multiple viewpoints provides an effective geometry-oriented measure of global 3D shape and completeness. A geometric error that may remain hidden in one projection, such as missing surfaces, excessive geometry, or an inaccurate head contour, is likely to be exposed from other viewpoints. Consequently, consistently high IoU across views indicates that the reconstruction agrees with the reference geometry from different viewing directions, substantially reducing the ambiguity of a single-view silhouette comparison. We evaluate all methods on both the AnyHead and NeRSemble datasets using the same camera views and evaluation protocol.

As shown in Table~\ref{tab:iou}, Any3DAvatar achieves the best multi-view 2D IoU on both datasets. Specifically, our method obtains an IoU of 0.9432 on AnyHead and 0.9060 on NeRSemble, outperforming all competing methods by clear margins. Since the metric jointly assesses silhouette agreement from multiple viewing directions, these improvements cannot be attributed to an isolated favorable projection. Instead, they demonstrate that Any3DAvatar maintains accurate geometric coverage across views and reconstructs more complete and consistent full-head geometry.

\begin{table*}[]
\centering
\small
\setlength{\tabcolsep}{4pt}
\begin{tabular}{lccccccc}
\hline
\textbf{Dataset} & \textbf{PanoHead} & \textbf{SphereHead} & \textbf{Arc2Avatar} & \textbf{FaceLift} & \textbf{ID-Sculpt} & \textbf{HQ-Head} & \textbf{Any3DAvatar} \\
\hline
\textbf{AnyHead} & 0.5299 & 0.5696 & 0.6502 & 0.8053 & 0.6799 & 0.8695 & \textbf{0.9432} \\
\textbf{NeRSemble} & 0.6297 & 0.6324 & 0.6135 & 0.6771 & 0.6090 & 0.6409 & \textbf{0.9060} \\
\hline
\end{tabular}%
\caption{Quantitative comparison using multi-view 2D IoU on the AnyHead and NeRSemble datasets.}
\label{tab:iou}
\end{table*}
\subsection{Comparison with Image-to-3D Methods}

We further compare Any3DAvatar against two recent state-of-the-art image-to-3D models, TRELLIS~\cite{trellis} and Open-Diffusion-GS~\cite{baking_gs}. To fairly measure full-head reconstruction quality, we conduct this comparison only on the AnyHead dataset. As shown in Fig.~\ref{fig:comparison_123d} and Table~\ref{tab:image_to_3d_comparison}, these general-purpose image-to-3D methods often produce unsatisfactory results for full-head reconstruction: they struggle to preserve complete identity information, and they also show limited understanding of full-head geometry and texture consistency, especially in non-frontal regions. However, Any3DAvatar consistently achieves better quantitative performance across all metrics and delivers stronger visual quality for complete head reconstruction.

\begin{figure}[t]
    \centering
    \includegraphics[width=\linewidth]{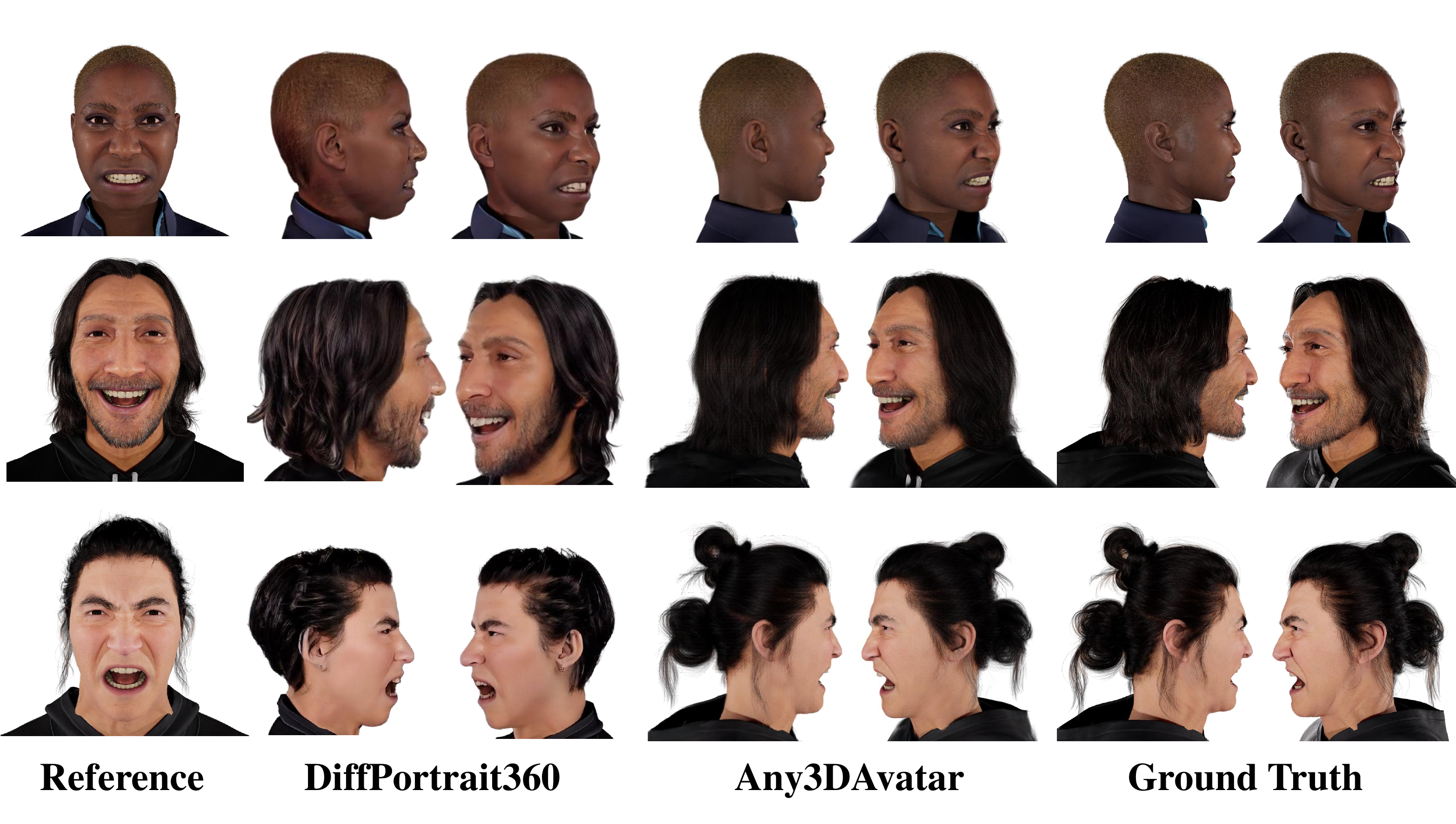}
    \caption{Visual comparison with non-full-head 3D head generation methods.}
    \label{fig:diffportrait}
\end{figure}

\begin{figure*}[t]
    \centering
    \includegraphics[width=\linewidth]{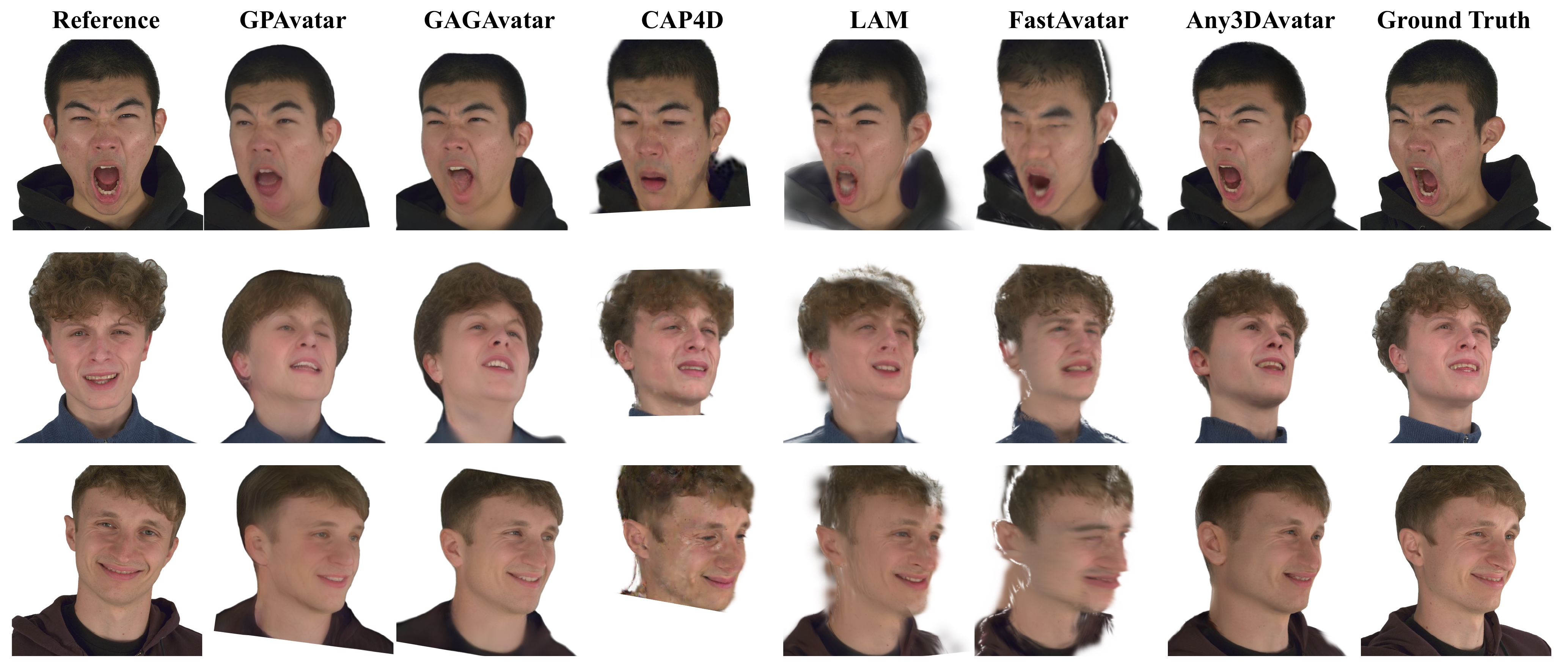}
    \caption{Visual comparison with non-full-head 3D head reconstruction methods.}
    \label{fig:4dhead}
\end{figure*}

\begin{table}[]
\centering
\small
\setlength{\tabcolsep}{2pt}
\begin{tabular}{cccccc}
\hline
\textbf{Method}      & \textbf{LPIPS} $\downarrow$ & \textbf{PSNR} $\uparrow$ & \textbf{SSIM} $\uparrow$ & \textbf{CSIM} $\uparrow$ & \textbf{DS} $\downarrow$ \\ \hline
\textbf{GPAvatar}    & 0.3233 & 12.2337 & 0.7652 & 0.8175  & 0.1410   \\
\textbf{GAGAvatar}   & 0.2968 & 12.8123 & 0.7727 & 0.8327  & 0.1149   \\
\textbf{CAP4D}       & 0.3297 & \underline{11.1998} & 0.7435 & 0.8251  & 0.1479   \\
\textbf{LAM}         & 0.3070 & 14.5441 & 0.7733 & 0.8098  & 0.1386   \\
\textbf{FastAvatar}  & \underline{0.3611} & 12.5209 & \underline{0.7328} & \underline{0.7312}  & \underline{0.1833}   \\
\textbf{Any3DAvatar} & \textbf{0.2357} & \textbf{16.6202} & \textbf{0.7988} & \textbf{0.8905}  & \textbf{0.0621}   \\ \hline
\end{tabular}
\caption{Quantitative comparison with non-full-head 3D head reconstruction methods on the NeRSemble dataset.}
\label{tab:none-full-head}
\end{table}

\subsection{Comparison with More Image-to-3D-Head Methods}

We note that single-image 3D head reconstruction methods are developed with slightly different goals, and thus are not always directly aligned with our full-head setting. In particular, some methods are designed to generate 360$^{\circ}$ turntable videos instead of recovering a canonical full-head representation for free-viewpoint rendering, while others mainly focus on reconstructing non-full-head regions rather than producing a complete full head. To provide a more comprehensive evaluation, we therefore first compare with 360$^{\circ}$ head turntable video generation methods and then with other non-full-head 3D head reconstruction methods.

\noindent\textbf{Comparison with 360$^{\circ}$ Head Turntable Video Generation Methods.}
We further compare our method with DiffPortrait360~\cite{diffportrait360} (CVPR 2025), a recent approach for generating 360$^{\circ}$ head turntable videos. Since DiffPortrait360 does not provide explicit camera control and is therefore not directly compatible with our evaluation protocol, we conduct a qualitative comparison on the AnyHead test set instead. As shown in Fig.~\ref{fig:diffportrait}, Any3DAvatar produces clearly better texture details and stronger multi-view consistency across different viewpoints. In contrast, although DiffPortrait360 benefits from an implicit representation, it still shows weaker consistency in head generation, leading to less stable geometry and appearance across views.

\noindent\textbf{Comparison with Other Non-Full-Head 3D Head Reconstruction Methods.}
We also compare Any3DAvatar with several representative methods that generate 3D heads from a single image but do not explicitly target full-head reconstruction. Specifically, we include GPAvatar~\cite{gpavatar} (ICLR 2024), GAGAvatar~\cite{gagavatar} (NeurIPS 2024), CAP4D~\cite{cap4d} (CVPR 2025), LAM~\cite{lam} (SIGGRAPH 2025), and FastAvatar~\cite{fastavatar} (ICLR 2026). Since these methods usually support only a limited range of viewpoints and are not designed for complete full-head recovery, we conduct the comparison on the NeRSemble~\cite{nersemble}, which better matches their evaluation setting. As shown in Table~\ref{tab:none-full-head} and Fig.~\ref{fig:diffportrait}, Any3DAvatar still achieves the best overall performance under this evaluation protocol, delivering the strongest quantitative results across all metrics as well as better visual quality. These results show that Any3DAvatar remains superior even on datasets and settings that are more favorable to non-full-head methods and restricted viewing ranges.

\subsection{Additional Baseline Training}
In the main paper, we evaluate all competing methods using the checkpoints released by their respective authors and follow the official inference settings whenever available. This protocol is necessary because most of the compared methods do not provide public training scripts, making it infeasible to retrain every baseline under a unified data setting. Among the evaluated baselines, FaceLift is the only method that releases an official training pipeline. We therefore conduct an additional experiment by training FaceLift on our AnyHead training dataset using its official implementation. We evaluate the retrained FaceLift and Any3DAvatar on the AnyHead dataset using the same evaluation protocol. This experiment controls for differences in training data and enables a more direct comparison between the two methods under the same data source.

\begin{figure}[t]
    \centering
    \includegraphics[width=\linewidth]{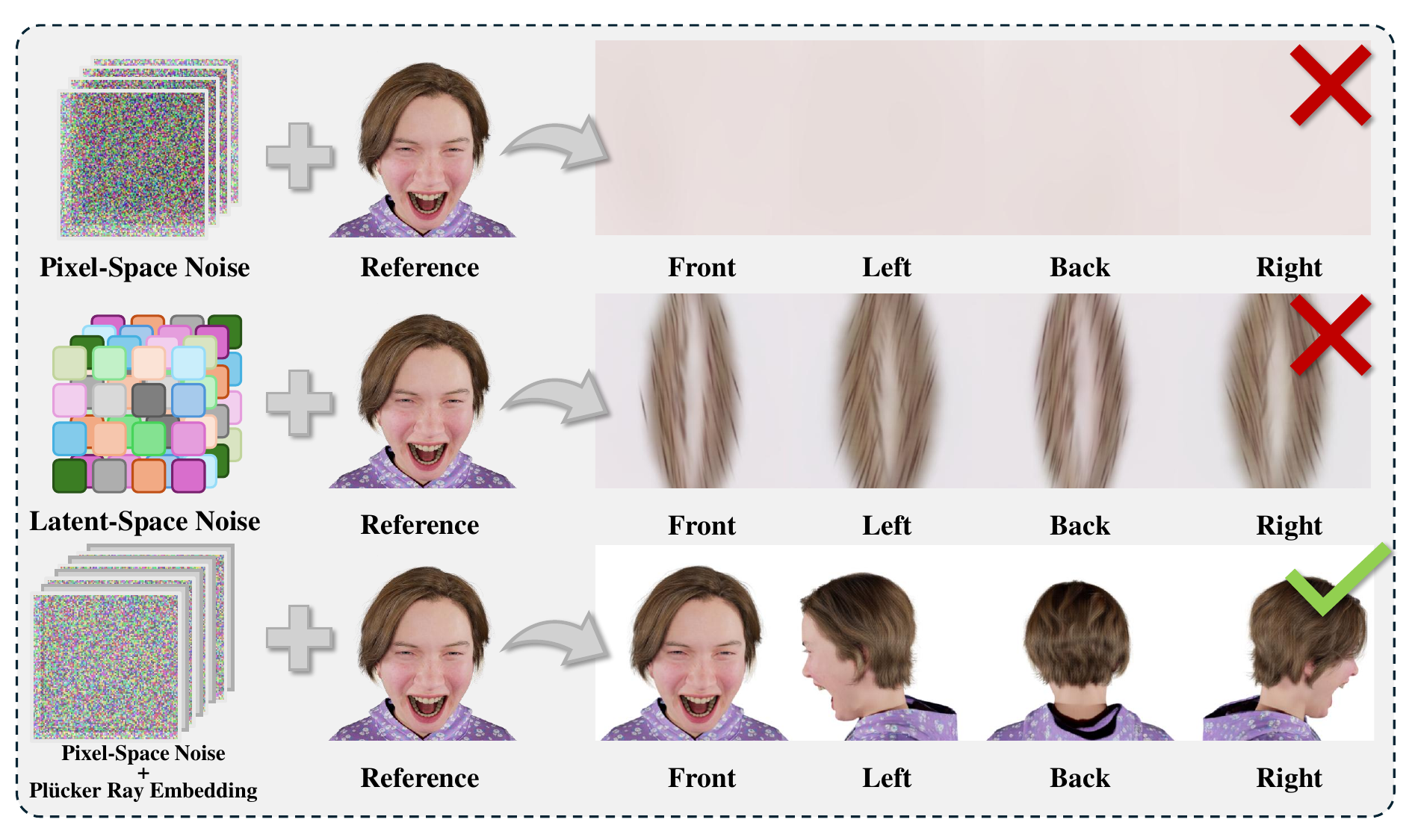}
    \caption{Training convergence of different noise initialization strategies. The variants using pure image-space Gaussian noise or direct latent-space noise fail to converge, whereas incorporating Pl\"ucker ray embedding enables stable convergence.}
    \label{fig:add_noise_position}
\end{figure}

\begin{table}[]
\centering
\small
\setlength{\tabcolsep}{2pt}
\begin{tabular}{cccccc}
\hline
\textbf{Method}           & \textbf{LPIPS}$\downarrow$  & \textbf{PSNR}$\uparrow$     & \textbf{SSIM}$\uparrow$    & \textbf{CSIM}$\uparrow$    & \textbf{DS}$\downarrow$     \\ \hline
\textbf{Trained FaceLift} & 0.2569          & 14.6061   & 0.7141          & 0.8711          & 0.1509          \\
\textbf{Any3DAvatar}      & \textbf{0.1894} & \textbf{19.8639} & \textbf{0.7632} & \textbf{0.9258} & \textbf{0.0633} \\ \hline
\end{tabular}
\caption{Quantitative comparison with FaceLift retrained and evaluated on the AnyHead dataset.}
\label{tab:facelift_training}
\end{table}

As shown in Table~\ref{tab:facelift_training}, when evaluated on the AnyHead dataset, Any3DAvatar consistently outperforms FaceLift across all evaluation metrics even after FaceLift is retrained on the same AnyHead training data. Compared with the retrained FaceLift, our method reduces LPIPS from 0.2569 to 0.1894 and DreamSim from 0.1509 to 0.0633, while improving PSNR from 14.6061 to 19.8639, SSIM from 0.7141 to 0.7632, and CSIM from 0.8711 to 0.9258. These results demonstrate that the performance advantage of Any3DAvatar does not arise solely from access to the AnyHead training data; the proposed model design also plays a crucial role in achieving higher reconstruction fidelity and identity consistency.

\begin{figure*}[t]
    \centering
    \includegraphics[width=\linewidth]{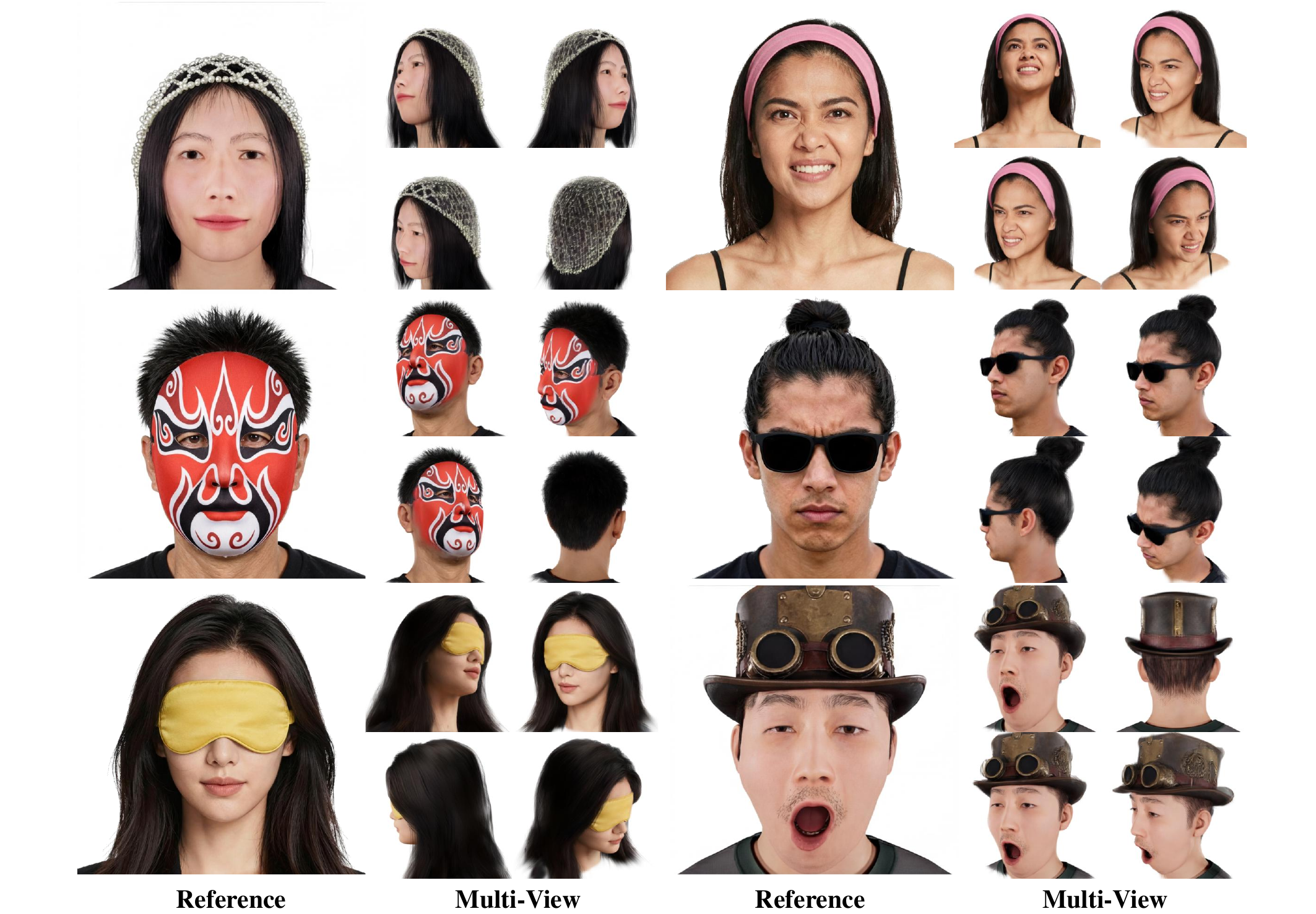}
    \caption{Visualized accessory rendering results. Our method provides diverse accessory generation capabilities.}
    \label{fig:accessory_case}
\end{figure*}

\begin{table}[]
\centering
\small
\setlength{\tabcolsep}{0.3mm}
\begin{tabular}{cccccc}
\hline
\textbf{Method}     & \textbf{LPIPS}$\downarrow$                      & \textbf{PSNR}$\uparrow$                & \textbf{SSIM}$\uparrow$                        & \textbf{CSIM}$\uparrow$                        & \textbf{DS}$\downarrow$                         \\ \hline
\textbf{Regression (Step=1)} & 0.2041                              & 18.6431            & 0.7531                              & 0.8954                              & 0.0693                              \\
\textbf{Endpoint Denoising (Step=1)} & 0.2038                              & 18.6689                     & 0.7541                              & 0.8970                              & 0.0690                              \\
\textbf{Endpoint Denoising (Step=5)} & \textbf{0.2014} & \textbf{18.6976}          & \textbf{0.7545} & \textbf{0.8998} & \textbf{0.0679} \\ \hline
\end{tabular}
\caption{Comparison of direct regression and conditional endpoint denoising under one- and five-step generation on the AnyHead dataset.}
\label{tab:regression}
\end{table}

\subsection{One-Step Generation: Regression vs. Conditional Endpoint Denoising}

One potential concern is that, when the number of denoising steps is set to one, our generation process may appear equivalent to a conventional regression-based method. However, using a single inference step does not make the two training paradigms equivalent. A regression model directly optimizes a one-step mapping to the reconstruction target, whereas our model uses a conditional endpoint denoising objective: it samples intermediate noisy states and learns to predict the clean endpoint conditioned on the input portrait and timestep. To investigate which training paradigm is more effective under the same one-step generation setting, we conduct a controlled comparison between direct regression and conditional endpoint denoising without AVAS. We keep the network architecture, training data, and one-step inference protocol unchanged, varying only the training objective. Both variants are trained and evaluated on the AnyHead dataset.

As shown in Table~\ref{tab:regression}, conditional endpoint denoising outperforms direct regression even when both methods use only one generation step. Compared with one-step regression, one-step endpoint denoising reduces LPIPS from 0.2041 to 0.2038 and DreamSim from 0.0693 to 0.0690, while improving PSNR from 18.6431 to 18.6689, SSIM from 0.7531 to 0.7541, and CSIM from 0.8954 to 0.8970. These consistent improvements across all metrics demonstrate that the advantage of our training strategy is not merely a consequence of multi-step inference; conditional endpoint denoising learns a better reconstruction function than direct one-step regression under the same inference budget. Moreover, increasing the number of endpoint-denoising steps from one to five further improves every metric and achieves the best overall performance. This result validates both the effectiveness of the conditional endpoint denoising objective and its ability to benefit from iterative refinement.
\begin{figure*}[t]
    \centering
    \includegraphics[width=\linewidth]{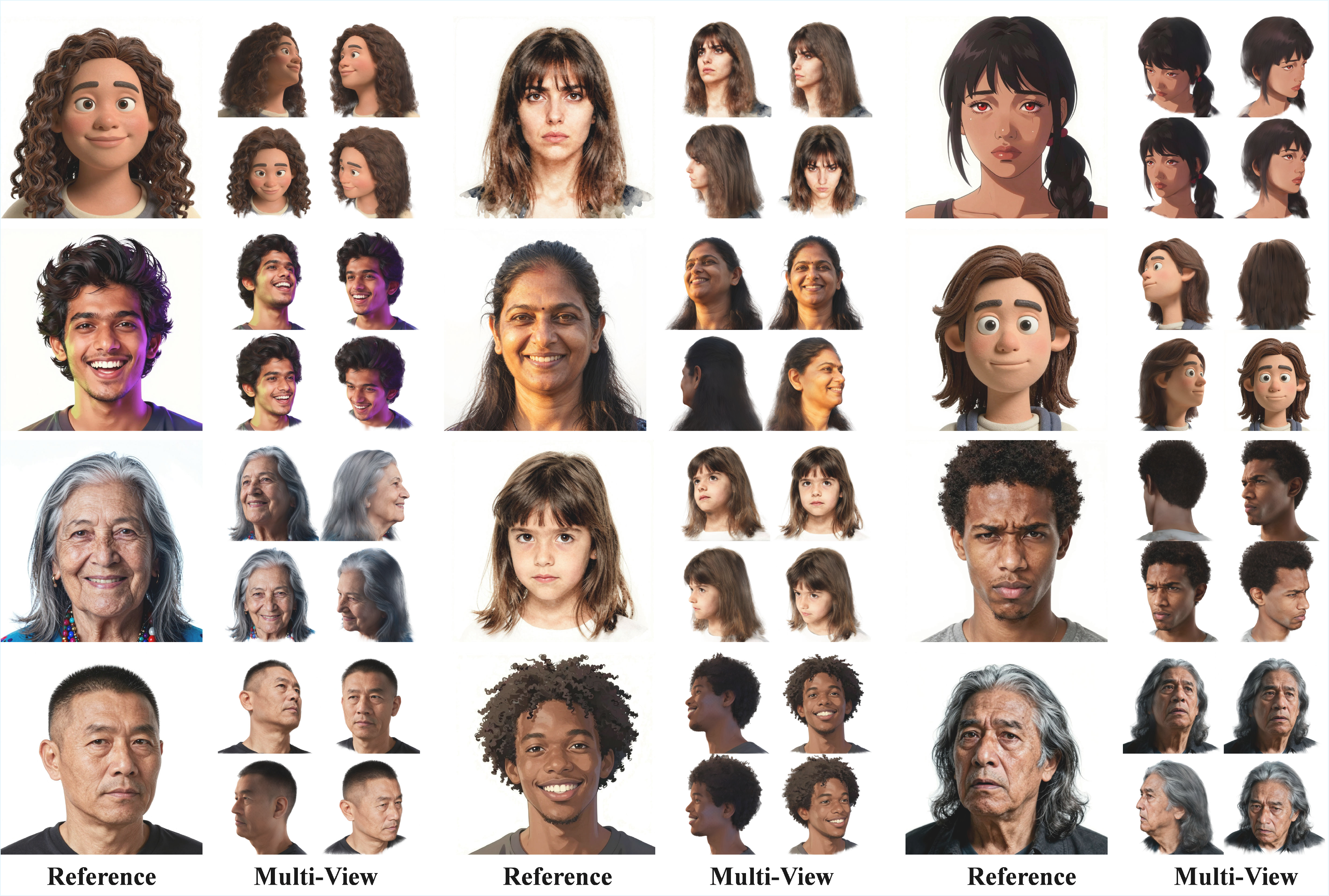}
    \caption{We evaluate our method on more stylized heads, including complex lighting, non-frontal viewpoints, anime-style heads, and 3D-stylized heads, demonstrating strong generation quality.}
    \label{fig:sytlized_results}
\end{figure*}

\subsection{Effect of Pl\"ucker Ray Embedding}

To investigate whether explicit ray information is necessary for learning 3D Gaussian reconstruction, we design three variants that differ in their noise representation and spatial encoding. First, we sample pure three-channel Gaussian noise in image space and pass it through the Gaussian encoder, testing whether the network can learn to reconstruct a 3D Gaussian point cloud from unstructured image noise alone. Second, we directly initialize the Gaussian tokens with random noise in the latent space, bypassing the image-space noise encoding process. Third, in our proposed design, we concatenate three-channel image-space Gaussian noise with a six-channel Pl\"ucker ray embedding and feed the resulting nine-channel tensor into the Gaussian encoder. In this way, every noise pixel is explicitly associated with its corresponding camera ray in 3D space.

As shown in Fig.~\ref{fig:add_noise_position}, neither pure image-space Gaussian noise nor direct latent-space noise provides sufficient spatial information for the model to establish a reliable mapping from noisy tokens to 3D Gaussian locations, causing both variants to fail to converge during training and preventing them from producing valid 3D Gaussian point clouds. In contrast, incorporating the Pl\"ucker ray embedding provides an explicit geometric prior that anchors each token to an oriented camera ray, leading to stable convergence and successful reconstruction. These results demonstrate that the Pl\"ucker ray embedding is essential for learning spatially coherent 3D Gaussian representations from noise.

\section{More Visual Results}

\subsection{Accessory Generation}

Leveraging the Accessory-Rich Heads data and our FLAME-free network design, our model gains strong accessory-aware generation capability. As shown in Fig.~\ref{fig:accessory_case}, Any3DAvatar can synthesize diverse accessories while preserving identity consistency and full-head structure. The visual results demonstrate high-quality generation with realistic geometry and texture details under challenging accessory variations. However, we also observe limitations: under some extreme viewpoints, accessory generation becomes less reliable and the generalization ability is still insufficient. We attribute this mainly to the limited scale and viewpoint coverage of the current accessory dataset. In future work, we will further expand dataset richness and improve model capacity for more robust accessory generation.

\subsection{More Stylized Visual Results}

We further evaluate Any3DAvatar on a broader range of head inputs, covering diverse ethnic groups, lighting conditions, stylization types, facial expressions, and viewing angles. As shown in Fig.~\ref{fig:sytlized_results}, the visual results indicate that our method consistently achieves high quality in multi-view consistency, identity preservation, and texture clarity across these challenging settings.

\section{Limitations and Future Work}

Despite the strong performance of Any3DAvatar, our current framework still has several limitations. First, for accessory-rich heads, the number of available viewpoints is still limited, which restricts both generation quality and generalization ability under challenging view changes. In future work, we plan to develop better data acquisition and processing pipelines to build large-scale multi-view-consistent accessory-rich head data. Second, the current model still contains a relatively large number of parameters, and its inference-time GPU memory consumption can reach around 10~GB. We therefore plan to release a more lightweight version in the future, with the goal of enabling efficient deployment on more devices. Third, the current dataset still covers only limited facial expressions and head poses. We will further extend the dataset with more diverse and complex expressions and poses, which is expected not only to improve current 3D head generation performance but also to provide a stronger foundation for future research on 4D head generation.

\end{document}